\documentclass{article}

     \PassOptionsToPackage{numbers, compress}{natbib}

    \usepackage[final]{neurips_2024}

\usepackage{amsmath,amsfonts,bm}

\def\eqref#1{equation~\ref{#1}}

\def\1{\bm{1}}

\DeclareMathAlphabet{\mathsfit}{\encodingdefault}{\sfdefault}{m}{sl}
\SetMathAlphabet{\mathsfit}{bold}{\encodingdefault}{\sfdefault}{bx}{n}

\usepackage{floatrow}
\newfloatcommand{capbtabbox}{table}[][\FBwidth]

\usepackage{blindtext}
\usepackage[utf8]{inputenc} 
\usepackage[T1]{fontenc}    
\usepackage[pdftex]{graphicx}
\usepackage[font=small]{caption}  
\usepackage{hyperref}       
\usepackage{url}            
\usepackage{booktabs}       
\usepackage{amsfonts}       
\usepackage{nicefrac}       
\usepackage{microtype}      
\usepackage{xcolor}         
\usepackage{wrapfig}
\usepackage{subcaption}
\usepackage{tabularx}
\usepackage{amsmath}
\usepackage{amssymb}
\usepackage{algorithm}
\usepackage{algpseudocode}
\usepackage{mathtools}
\usepackage{copyrightbox}
\usepackage{bbm}
\usepackage{amsthm}
\usepackage{chngcntr}
\usepackage{enumitem}
\usepackage{xspace}
\usepackage{comment}
\usepackage{placeins}
\usepackage{titlesec}
\usepackage{rotating}
\usepackage{multirow}
\usepackage{colortbl}
\usepackage{longtable}
\usepackage{multicol}
\usepackage[para]{footmisc}
\usepackage{hyperref}
\titlespacing{\paragraph}{0pt}{0pt}{0.5em}
\titlespacing{\section}{0pt}{0pt}{0pt}
\titlespacing{\subsection}{0pt}{0pt}{0pt}
\definecolor{mygray}{gray}{0.93}

\title{When Does Perceptual Alignment \\Benefit Vision Representations?}

\author{
Shobhita Sundaram$^{1}$\thanks{Equal contribution.}
\hspace{5mm}
Stephanie Fu$^{2*}$
\hspace{5mm}
\textbf{Lukas Muttenthaler}$^{3,4}$\thanks{Work partly done while a Student Researcher at Google DeepMind.}
\vspace{3mm} \\
\textbf{Netanel Y. Tamir}$^{5}$
\hspace{4mm}
\textbf{Lucy Chai}$^{1}$
\hspace{4mm}
\textbf{Simon Kornblith}$^{6}$
\hspace{4mm}
\textbf{Trevor Darrell}$^{2}$
\hspace{4mm}
\textbf{Phillip Isola}$^{1}$
\vspace{3mm}
\\
$^{1}$MIT \hspace{2mm} $^{2}$U.C. Berkeley \hspace{2mm} $^{3}$TU Berlin \hspace{2mm} $^{4}$BIFOLD \hspace{2mm} $^{5}$Weizmann Institute of Science \hspace{2mm} $^{6}$Anthropic
}

\begin{document}

\maketitle

\vspace{-12mm}
\begin{figure*}[ht!]
    \centering
    \includegraphics[width=.8\linewidth]{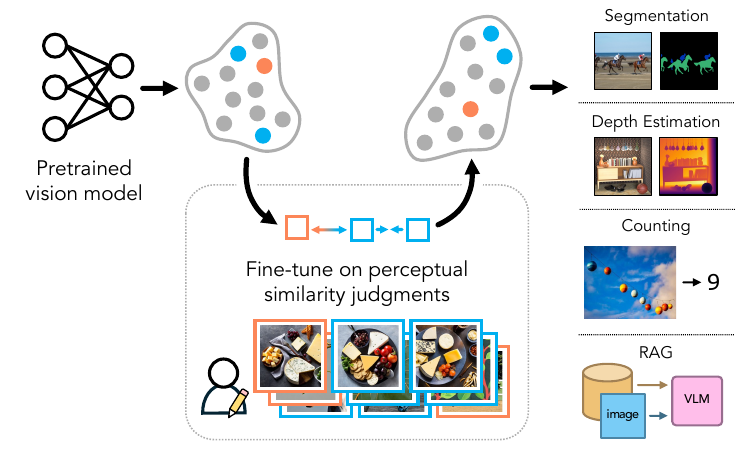}
    \caption{\small \textbf{Does human perceptual alignment improve vision representations?} Vision models have been shown to learn useful image representations through large-scale pretraining (e.g., CLIP, DINO). We find that additionally aligning these models to human perceptual judgments yields representations that \emph{improve} upon the original backbones across many downstream tasks, including counting, segmentation, depth estimation, instance retrieval, and retrieval-augmented generation. Our blog post and code are available at \url{percep-align.github.io}.}
    \label{fig:teaser}
\end{figure*}
\vspace{-2mm}
\begin{abstract}
\vspace{-3mm}
Humans judge perceptual similarity according to diverse visual attributes, including scene layout, subject location, and camera pose. Existing vision models understand a wide range of semantic abstractions but improperly weigh these attributes and thus make inferences misaligned with human perception. 
While vision representations have previously benefited from alignment in contexts like image generation, the utility of perceptually aligned representations in more general-purpose settings remains unclear. Here, we investigate how aligning vision model representations to human perceptual judgments impacts their usability across diverse computer vision tasks. We finetune state-of-the-art models on human similarity judgments for image triplets and evaluate them across standard vision benchmarks. We find that aligning models to perceptual judgments yields representations that \emph{improve} upon the original backbones across many downstream tasks, including counting, segmentation, depth estimation, instance retrieval, and retrieval-augmented generation. In addition, we find that performance is widely preserved on other tasks, including specialized \emph{out-of-distribution} domains such as in medical imaging and 3D environment frames. Our results suggest that injecting an inductive bias about human perceptual knowledge into vision models can contribute to better representations.
\end{abstract}

\section{Introduction}
\vspace{-1mm}
Our sense of similarity is crucial to how we perceive and act in the world. Consider the range of factors that might influence your judgment of visual similarity: layout, color, perspective, semantics, and more. These characteristics shape our local inferences about object relationships, enabling us to build a comprehensive understanding of the visual world.

Given the importance of similarity judgments in our visual perception, it follows that aligning vision models to these judgments could help them develop more human-like visual capabilities. 
In the language domain, we have seen the power of aligning LLMs to human feedback~\citep[RLHF;][]{rlhp2017}, resulting in safer and more useful models~\cite{bai2022training}.
Similar trends are emerging in vision, where there is growing interest in alignment with human perceptual judgments to improve vision representations~\citep[e.g.,][]{lpips, NEURIPS2022_harmonizing, muttenthaler2023improving, dreamsim, muttenthaler2024aligning}. Human feedback has also been used to improve the aesthetic quality of diffusion-generated images~\citep{hps_2023,fan2023dpok, kirstain2023pickapic}, increase retrieval capabilities~\citep{dreamsim}, and improve downstream task performance of image/text models~\citep{muttenthaler2023improving}. While acknowledging this wide space of human preference data used for alignment in vision and language, we focus our study on image similarity judgments, which give us a more stable and shared measure of human visual perception ~\citep{dreamsim, muttenthaler2023human, sucholutsky2023getting, muttenthaler2024aligning}.
    
    Although there is consensus that alignment to perceptual judgements can enable specific goals (e.g., similarity \cite{dreamsim, lpips, muttenthaler2023human, muttenthaler2023improving}), their utility in creating \emph{general-purpose} representations is less clear. 
    Does human alignment improve model performance by leveraging human-provided labels as an inductive bias, or compromise the model's original representational power by diverting it towards a separate task? Recent work suggests that the conclusions are nuanced: naively incorporating human perceptual knowledge into model finetuning can distort representations, requiring strong regularization to maintain downstream performance while increasing the representations' interpretability and alignment~\cite{muttenthaler2023improving}. Furthermore, objective function and training data appear to matter more than architecture or model size for aligning to perceptual judgments~\cite{muttenthaler2023human}. The nature of human preference labels also plays a crucial role, as different types of labels produce distinct learning signals. For instance, fine-tuning vision backbones with mid-level perceptual similarity judgments yields representations well-suited for image retrieval tasks~\cite{dreamsim}, while low-level similarity labels are more effective for image reconstruction losses~\cite{lpips}.

    It is clear from this body of work that careful perceptual alignment helps with specific tasks, however the effect of these adjustments on the models' representation spaces is less well understood. That is, are models aligned to perceptual judgments only better at specific tasks -- such as predicting image similarity -- or do they, like humans, actually have better general-purpose representations? 
    Here, we evaluate the usefulness of human-aligned representations not just in predicting perceptual judgments but also on standard vision benchmarks requiring diverse notions of visual understanding. 
    We finetune several state-of-the-art models -- including CLIP, DINO, DINOv2, and SynCLR -- on NIGHTS, a dataset of human similarity judgments over synthetic image triplets \citep{dreamsim}, and evaluate the finetuned models on standard tasks. 
    Our experiments suggest that these human-aligned representations demonstrate strong gains compared to the original model backbones, even on tasks requiring skills beyond those needed during training or perceptual alignment. 
    We also identify limitations of human perceptual alignment, and find that finetuning can sacrifice performance on some natural data tasks in which models had strong prior performance. 
    \vspace{-1mm}
    In summary, our contributions are the following:
    \begin{itemize}
        \item We investigate the effects of aligning pretrained vision models to human perceptual judgments on various downstream visual recognition tasks.
        \item We find that propagating global image-level human similarity annotations to ViT patch tokens benefits downstream dense prediction tasks such as depth prediction and segmentation.
        \item We show that human-aligned representations are also beneficial in retrieval-based tasks requiring global image understanding, including retrieval-augmented generation for recent vision-language models, counting-based retrieval, and instance-based retrieval. 
        \item We ablate the effect of various human-annotated visual similarity datasets, and find that mid-level image similarity, as opposed to low-level pixel-level variations or high-level semantic associations, offers the largest improvements in generalization capabilities. 
    \end{itemize}
\section{Related Works}
\label{sec:rel_works}

\paragraph{Vision backbones as learned feature extractors.}

Using the intermediate activations of deep networks has been a long-standing strategy, originally used as a way of harnessing data-driven priors from models trained on fewer large datasets and repurposing them for downstream tasks where such expensive supervision is less readily available \cite{donahue2014decaf,sharif2014cnn}. For instance, ImageNet pretrained models have proven useful for tasks such as texture synthesis \citep{gatys2015texture}, style-transfer \citep{gatys2015neural}, and super-resolution \citep{johnson2016perceptual}. But given the challenges of collecting large-scale labelled datasets using human annotators, other approaches such as self-supervised learning \citep{simclr, simclrv2, barlowtwins, DINO_2021, oquab2023dinov2} and contrastive learning on image/alt text pairs \citep{clip-models,openclip,align} have since superseded supervised learning as the standard approach for training vision models. Such models have been shown to yield rich multipurpose representations that can generalize to a variety of tasks involving visual perception, scene understanding, and image generation~\cite{sharma2023materialistic,liang2023semantic,shen2023distilled,kobayashi2022decomposing,luo2024diffusion,tumanyan2024dino}.

\paragraph{View selection for self-supervised learning.}

The self-supervised, contrastive-learning objective aims to maximize the feature similarity between similar views of the data and minimize the similarity between different (negative) views. While originally the negative views are selected randomly from a pool of images~\cite{simclr, moco}, this often results in requiring large batch sizes to learn useful representations. The process of selecting these positive and negative learning examples remains an active research area.  Recent approaches have suggested that alternative strategies including hard-negative mining~\cite{robinson2020contrastive}, nearest neighbors for positive pairs \cite{dwibedi2021little}, and supervised labels when available~\cite{khosla2020supervised} can provide useful learning signals. Other avenues include training the contrastive framework on synthetic data from generative models~\cite{jahanian2021generative, Tian2023LearningVF}, which provides an infinite data source of image variations rather than the traditional data augmentations typically used to generate alternative views.

\paragraph{Learning with human alignment.}

Learning directly from human feedback can provide models with more targeted supervision using fewer examples~\cite{rlhp2017}, and, thus, has been beneficial for fine-tuning large models towards specific human preferences~\cite{sohn2023styledrop,zhang2023hive}. \citet{ding2023quality} show that image similarity metrics, trained on human feedback, can be subsequently used for evaluating the diversity of a text-to-image model. Several datasets aim to annotate human visual preferences in image similarity, including low-level~\cite{lpips} and high-level~\cite{things-concepts} image variations. In particular, \citet{lpips} and \citet{dreamsim} use these datasets to learn a more human-aligned perceptual metric that improves the image retrieval abilities of vision models, while \citet{muttenthaler2023improving} used THINGS for learning a linear transform on top of vision representations to increase downstream task performance and improve alignment with human similarity judgments. These annotated human preference datasets serve as useful signals for contrastive objectives, providing direct supervision for positive and negative pairings that are aligned with human decisions; here, we investigate how fine-tuning self-supervised models using these human annotations impacts model performance on various downstream tasks.

\section{Learning from perceptual judgments}
We propose to use the method described below as a "second pretraining stage", which aligns the feature representations from large vision models with human perceptual judgments before applying them to downstream tasks. We note that prior work on this dataset aimed to develop a model for measuring image similarity based on human judgments. Here, we investigate if pretraining on this dataset leads to a better general-purpose representation, as measured by performance on different downstream tasks.

\subsection{Human Similarity Annotations}
We use the NIGHTS dataset to produce human-aligned variations of several large vision models~\cite{dreamsim}. The NIGHTS dataset consists of 20k synthetically generated image triplets, annotated with two alternative forced-choice human similarity judgments. These triplets are collected so that each has 6-10 unanimous human ratings, thus eliminating ambiguous cases where humans are likely to disagree. 

NIGHTS consists of image triplets varying in \textit{mid-level} information. Images in a triplet roughly share the same semantic content; however, they vary in pose, layout, shape, color, and the number of objects (see Fig.~\ref{fig:dataset} in the Appendix for examples). Thus, the perceptual judgments indicate the shared visual appearance properties, as opposed to requiring higher-level semantic knowledge about the image content.

\subsection{Image-level objective}
\label{subsec:objective}
Given a pre-trained backbone $f_{\theta}$, we fine-tune its parameters $\theta$ on a dataset of triplets $\mathcal{D}=\{(x, \tilde{x_0}, \tilde{x_1}), y\}$, where $x$ denotes a reference image, and $\tilde{x_0}$ and $\tilde{x_1}$ denote two variation images. The judgement $y \in \{0,1\}$ indicates which of $\tilde{x_0}$ and $\tilde{x_1}$ is more similar to $x$. We measure distance (dissimilarity) between two images $(x,\tilde{x_0})$ using the cosine distance between their respective image features $(f_{\theta}(x),f_{\theta}(\tilde{x_0}))$, which is defined as:
\begin{equation}
    d(x, \tilde{x_0}) = 1 - \frac{f_{\theta}(x) \cdot f_{\theta}(\tilde{x_0})}{|f_{\theta}(x)||f_{\theta}(\tilde{x_0})|}.
\end{equation}
We use an alignment loss to encourage the model to match human preferences:
\begin{equation}
    \mathcal{L}_{\text{alignment}}(\theta) = \max(0, m-\Delta d \cdot \bar{y}),
\label{eq:alignment_loss}
\end{equation}
where $\Delta d = d(x, \tilde{x_0}) - d(x, \tilde{x_1})$, $\bar{y}$ maps $y \in \{0, 1\} \rightarrow \{-1, 1\}$, and $m$ is the margin, which we set to 0.05 following \cite{dreamsim}. Note that this loss is equivalent to the triplet loss \citep{chechik2010large}, and thus minimizes the cosine distance between the representations of the more similar pair and maximizes the distance between the representations of the other pair. 

\subsection{Patch-level objective}
\label{subsec:patch}
\begin{figure*}[ht!]
    \centering
    \includegraphics[width=\linewidth]{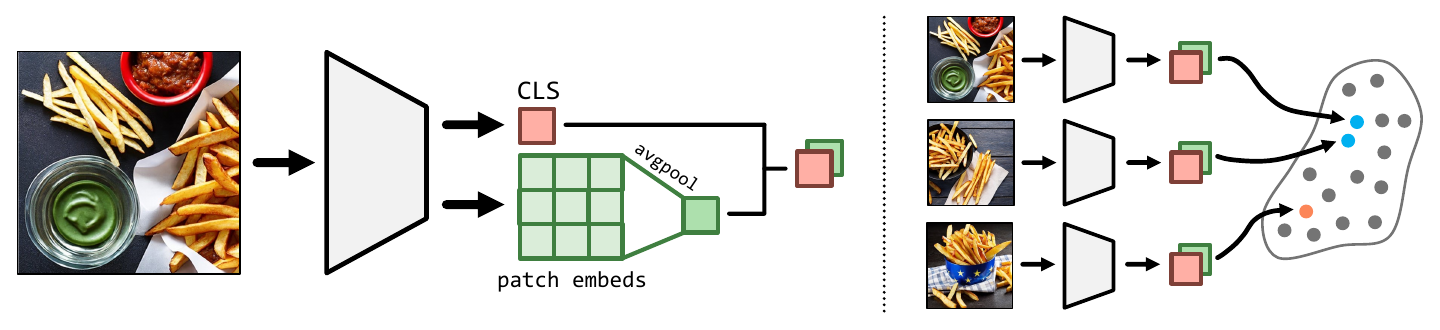}
    \caption{\small \textbf{Diagram of our feature extraction method when training with a patch-level objective.} Left: We extract the CLS and patch embeddings from DINO and DINOv2, perform a spatial average-pool on the patch embeddings, and concatenate [CLS, patch] vectors. Right: We train these concatenated features with a hinge loss, identical to the image-level objective.}
    \label{fig:patch}
\end{figure*}

The NIGHTS dataset contains global annotations of similarity at the image level, but the holistic label is the result of several local attributes, such as perspective, layout, foreground appearance etc. In addition to the global $\mathrm{CLS}$ token from the Vision Transformer model backbones, each model also contains a set of spatial patch embeddings. Propagating this global human annotation to individual patch tokens allows for spatial representations that are aligned with human similarity preferences. Thus, we formulate a patch alignment objective to optimize these local patch features jointly with the global image label. 

The local objective only differs from the global objective in how the features are extracted. Instead of computing $\mathcal{L}\left(\mathrm{CLS_A},\mathrm{CLS_B}\right)$, we compute $$\mathcal{L}\left(\mathrm{cat[CLS_A, pool(PATCH_A)], cat[CLS_B, pool(PATCH_B)]}\right).$$ $\mathrm{CLS}$ is of dimension $(1, d)$ and $\mathrm{PATCH}$ is $(s, s, d)$ where $s$ is the number of patches along each spatial dimension. We spatially average the patch tokens to get dimension $(1, d)$. We then concatenate the $\mathrm{CLS}$ and pooled patch tokens to get dimension $(1, 2d)$. We fine-tune the same alignment loss (see Eq.~\ref{eq:alignment_loss}) on concatenated $\mathrm{CLS}$ and averaged-pooled patch tokens, and train heads for semantic segmentation and depth estimation on the resulting patch embeddings. Note that only experiments reported in Sec.~\ref{sec:dense_pred} use this objective, as they require local features. For all other evaluations, we exclusively use $\mathrm{CLS}$ tokens (see Eq.~\ref{eq:alignment_loss}) if not mentioned otherwise.

\subsection{Implementation details}

\paragraph{Vision Model Backbones.} We fine-tune several state-of-the-art Vision Transformer \citep[VIT;][]{vit} backbones including DINO \cite{DINO_2021}, DINOv2 \cite{oquab2023dinov2}, CLIP \cite{clip-models}, OpenCLIP \cite{openclip}, and SynCLR \cite{Tian2023LearningVF}. For DINO, DINOv2, MAE, and SynCLR, we use the \texttt{CLS} token of the final layer. We extract the \texttt{CLS} token before layer norm for MAE and after it for the other backbones. For CLIP and OpenCLIP, we use the representations of the image encoder. We also experiment with concatenated features from DINO, CLIP, and OpenCLIP, used to train the DreamSim Ensemble in \cite{dreamsim} (referred to as "ensemble" in our results). For each model, we use its base size, ViT-B.

We implicitly ablate the usage of synthetic triplets in NIGHTS by including SynCLR in our experiments; as that backbone was contrastively trained on generated images, additional performance changes can be attributed to human perceptual alignment.

\paragraph{Finetuning with human preference labels.} We finetune each backbone using Low-Rank Adaptation (LoRA), which was found to achieve better alignment performance and efficiency than full fine-tuning in \cite{dreamsim}. For more training and technical details see Sections \ref{ha-details} and \ref{compute-details} in the Appendix. 

\section{Experiments}
\label{sec:experiments}

In this section, we evaluate perceptually-aligned backbones against base models on common vision tasks. We study global representations through instance retrieval, object-counting, and retrieval-augmented generation experiments. Additionally, we find that local patch-level representations can be improved by tuning on image-level perceptual judgments, and show performance increases on semantic segmentation and depth estimation.

\subsection{Dense Prediction}
\label{sec:dense_pred}

\noindent{\bf Semantic segmentation.} Following the procedure detailed in Section \ref{subsec:patch} and Fig.~ \ref{fig:patch}, we LoRA-tune new backbones with perceptually-aligned CLS and patch tokens. To evaluate segmentation performance, we freeze these backbones and train a single linear layer transforming patch tokens to a segmentation map. We evaluate DINO and DINOv2 on standard segmentation benchmarks in Table \ref{tab:seg} and show that human-aligned models boost performance in 16 out of 20 cases. Across all datasets and metrics, human-aligned DINO (denoted as DINO-HA) outperforms the base model and often achieves the highest mIoU and Pixel Accuracy (P.A.) overall. DINOv2-HA also outperforms its nonaligned counterpart on COCO and DAVIS2017.

We flag datasets already seen in the DINOv2 retrieval pretraining \cite{oquab2023dinov2}, such as Pascal VOC, ADE20k, and Cityscapes. If a dataset is already in-distribution for a backbone, fine-tuning on different data may be more likely to change the feature space such that that dataset is more out-of-distribution. Thus this is a potential confounding factor.

\begin{table*}[h]
\begin{center}
\resizebox{0.95\linewidth}{!}{
\setlength{\aboverulesep}{0.5pt}
\setlength{\belowrulesep}{0.5pt}
\setlength{\extrarowheight}{.75ex}
\begin{tabular}{l
>{\columncolor{mygray}}c
>{\columncolor{mygray}}ccc
>{\columncolor{mygray}}c
>{\columncolor{mygray}}ccc
>{\columncolor{mygray}}c
>{\columncolor{mygray}}c}

\toprule
\multicolumn{1}{l}{\textbf{}} & \multicolumn{2}{c}{\cellcolor{mygray}\bf Pascal VOC$^*$} & \multicolumn{2}{c}{\bf ADE20k$^*$} & \multicolumn{2}{c}{\cellcolor{mygray}\bf Cityscapes$^*$} & \multicolumn{2}{c}{\bf COCO} & \multicolumn{2}{c}{\cellcolor{mygray}\bf DAVIS2017} \\
\multicolumn{1}{l}{} & \multicolumn{1}{c}{\cellcolor{mygray}mIoU } & \multicolumn{1}{c}{\cellcolor{mygray}P.A. } & \multicolumn{1}{c}{mIoU } & \multicolumn{1}{c}{P.A. } & \multicolumn{1}{c}{\cellcolor{mygray}mIoU } & \multicolumn{1}{c}{\cellcolor{mygray}P.A. } & \multicolumn{1}{c}{mIoU } & \multicolumn{1}{c}{P.A. } & \multicolumn{1}{c}{\cellcolor{mygray}mIoU } & \multicolumn{1}{c}{\cellcolor{mygray}P.A. } \\
\midrule
DINO & 0.729 & 0.800 & 0.342 & 0.548 & 0.562 & 0.749 & 0.810 & 0.901 & 0.809 & 0.886 \\
DINO-HA & \textbf{0.745}$^{\dagger}$ & \textbf{0.840}$^{\dagger}$ & \textbf{0.391} & \textbf{0.602} & \textbf{0.573}$^{\dagger}$ & \textbf{0.769}$^{\dagger}$ & \textbf{0.816}$^{\dagger}$ & \textbf{0.914}$^{\dagger}$ & \textbf{0.818}$^{\dagger}$ & \textbf{0.913}$^{\dagger}$ \\
\midrule
DINOv2 & \textbf{0.686} & \textbf{0.803} & \textbf{0.441}$^{\dagger}$ & 0.645 & \textbf{0.535} & 0.737 & 0.759 & 0.877 & 0.794 & 0.906 \\
DINOv2-HA & 0.635 & 0.771 & 0.418 & \textbf{0.700}$^{\dagger}$ & 0.528 & \textbf{0.739} & \textbf{0.761} & \textbf{0.891} & \textbf{0.810} & \textbf{0.908} \\ \bottomrule
\end{tabular}
}
\end{center}
\caption{\small \textbf{Base and human-aligned model performance on semantic segmentation.} Aligned models largely outperform baselines, with DINO-HA achieving the highest performance across models for 4 out of 5 datasets. Note that Pascal VOC, ADE20k, and Cityscapes were included in DINOv2's retrieval pretraining. $\dagger$ indicates best score in the column.} 
\label{tab:seg}
\end{table*}
\vspace{-3mm}
\noindent{\bf Depth estimation.} We follow the evaluation protocol of \cite{oquab2023dinov2, li2022binsformer} and train a single linear layer on frozen patch tokens to output a depth map with values mapped into 256 uniformly-distributed bins. This head is trained with a scale-invariant log loss introduced in \cite{eigen2014depth} and a scale-invariant gradient-matching term as described in \cite{MegaDepthLi18}. In Table \ref{tab:depth}, we report performance on monocular depth estimation and show that human-aligned models outperform base models in 27 out of 36 cases. Consistent with segmentation performance, human-aligned DINO outperforms the base model on all metrics across all datasets, and is often the highest-performing model overall (denoted by $\dagger$). We also evaluate out-of-distribution generalization by training a depth head on NYUv2 and evaluating on the 4D Light Field dataset; combined with the performance boost even on datasets that DINOv2 was trained on (NYUv2 and SUN-RGBD), these results demonstrate that human-aligned models have strong generalization capabilities prior to any training for downstream tasks.

\begin{table*}[ht!]
\begin{center}
\resizebox{0.9\linewidth}{!}{
\setlength{\aboverulesep}{0.5pt}
\setlength{\belowrulesep}{0.5pt}
\setlength{\extrarowheight}{.75ex}
\begin{tabular}{l
>{\columncolor{mygray}}cc
>{\columncolor{mygray}}cc
>{\columncolor{mygray}}cc}
\toprule
\multicolumn{1}{l}{\textbf{}} & \multicolumn{6}{c}{\bf NYUv2$^*$} \\
\multicolumn{1}{l}{} & \multicolumn{1}{l}{\cellcolor{mygray}RMSE ($\downarrow$)} & \multicolumn{1}{l}{AbsRel ($\downarrow$)} & \multicolumn{1}{l}{\cellcolor{mygray}log10 ($\downarrow$)} & \multicolumn{1}{l}{$\delta > 1.25$ ($\uparrow$)} & \multicolumn{1}{l}{\cellcolor{mygray}$\delta > 1.25^2$ ($\uparrow$)} & \multicolumn{1}{l}{$\delta > 1.25^3$ ($\uparrow$)} \\
\midrule
DINO & 1.034 & \textbf{3.517} & 0.173 & 0.415 & 0.746 & 0.895 \\
DINO-HA & \textbf{1.032} & 3.759 & \textbf{0.169}$^{\dagger}$ & \textbf{0.445}$^{\dagger}$ & \textbf{0.761} & \textbf{0.900} \\
\midrule
DINOv2 & \textbf{1.003}$^{\dagger}$ & 3.188 & \textbf{0.178} & \textbf{0.445}$^{\dagger}$ & \textbf{0.785}$^{\dagger}$ & \textbf{0.907}$^{\dagger}$ \\
DINOv2-HA & 1.062 & \textbf{3.167}$^{\dagger}$ & 0.185 & 0.419 & 0.749 & 0.891 \\ 
\midrule
\multicolumn{1}{l}{\textbf{}} & \multicolumn{6}{c}{\bf NYUv2 $\rightarrow$ 4D LF} \\
\multicolumn{1}{l}{} & \multicolumn{1}{l}{\cellcolor{mygray}RMSE ($\downarrow$)} & \multicolumn{1}{l}{AbsRel ($\downarrow$)} & \multicolumn{1}{l}{\cellcolor{mygray}log10 ($\downarrow$)} & \multicolumn{1}{l}{$\delta > 1.25$ ($\uparrow$)} & \multicolumn{1}{l}{\cellcolor{mygray}$\delta > 1.25^2$ ($\uparrow$)} & \multicolumn{1}{l}{$\delta > 1.25^3$ ($\uparrow$)} \\
\midrule
DINO & 3.817 & 0.543 & 0.380 & 0.160 & 0.335 & 0.454 \\
DINO-HA & \textbf{3.757} & \textbf{0.538} & \textbf{0.362} & \textbf{0.187} & \textbf{0.346} & \textbf{0.463} \\
\midrule
DINOv2 & \textbf{3.460}$^{\dagger}$ & 0.497 & \textbf{0.337}$^{\dagger}$ & 0.194 & 0.337 & \textbf{0.464}$^{\dagger}$ \\
DINOv2-HA & 3.720 & \textbf{0.487}$^{\dagger}$ & 0.356 & \textbf{0.203}$^{\dagger}$ & \textbf{0.356}$^{\dagger}$ & 0.455 \\ 
\midrule
\multicolumn{1}{l}{\textbf{}} & \multicolumn{6}{c}{\bf SUN-RGBD$^*$} \\
\multicolumn{1}{l}{} & \multicolumn{1}{l}{\cellcolor{mygray}RMSE ($\downarrow$)} & \multicolumn{1}{l}{AbsRel ($\downarrow$)} & \multicolumn{1}{l}{\cellcolor{mygray}log10 ($\downarrow$)} & \multicolumn{1}{l}{$\delta > 1.25$ ($\uparrow$)} & \multicolumn{1}{l}{\cellcolor{mygray}$\delta > 1.25^2$ ($\uparrow$)} & \multicolumn{1}{l}{$\delta > 1.25^3$ ($\uparrow$)} \\
\midrule
DINO & 4.900 & 2.350 & 0.533 & 0.205 & 0.385 & 0.524 \\
DINO-HA & \textbf{4.788}$^{\dagger}$ & \textbf{2.300}$^{\dagger}$ & \textbf{0.526}$^{\dagger}$ & \textbf{0.230} & \textbf{0.418}$^{\dagger}$ & \textbf{0.531}$^{\dagger}$ \\
\midrule
DINOv2 & 5.200 & 3.023 & 0.615 & 0.173 & 0.309 & 0.444 \\
DINOv2-HA & \textbf{5.082} & \textbf{2.904} & \textbf{0.599} & \textbf{0.237}$^{\dagger}$ & \textbf{0.409} & \textbf{0.487} \\ 
\bottomrule
\end{tabular}
}
\end{center}
\caption{\small{\textbf{Human-aligned DINO and DINOv2 performance on monocular depth estimation benchmarks.} Note that NYUv2 and SUN-RGBD were included in DINOv2's retrieval pretraining set, yet human-aligned DINOv2 still outperforms the base model on SUN-RGBD. Along with the results on an unseen test data domain (train on NYUv2 $\rightarrow$ test on 4D Light Field), these results demonstrate strong generalization performance of models aligned to human perceptual judgments. $\dagger$ indicates best score in the column.}}
\label{tab:depth}
\end{table*}

\subsection{Retrieval-augmented generation}
\label{sec:rag}
\begin{figure*}[ht!]
    \centering    \includegraphics[width=\linewidth]{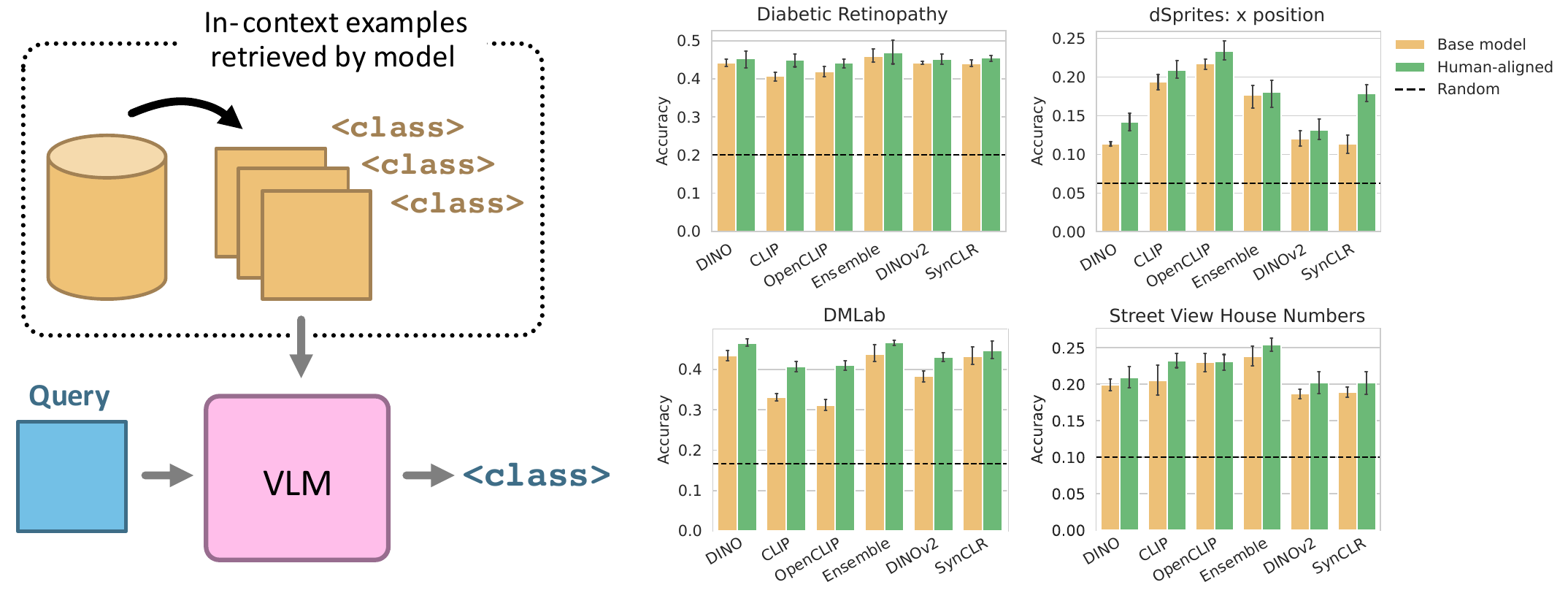}
    \caption{\small \textbf{Left: Diagram of evaluation setup for retrieval-augmented generation.} We retrieve the top-3 nearest image-prompt examples for each datasets and prompt OpenFlamingo with them before inputting the query image. \textbf{Right: Classification accuracy on VTAB \cite{li2019large} from wide-varying domains.} Error bars indicate 95\% confidence interval over 5 random seeds.}
    \label{fig:rag}
\end{figure*}
First introduced for text generation with non-parametric memory \cite{lewis2021retrievalaugmented}, retrieval-augmented generation (RAG) has become a popular method for selecting relevant few-shot examples when prompting large vision-language models (VLMs) \cite{alayrac2022flamingo, awadalla2023openflamingo, laurencon2023obelics, laurençon2024matters}. RAG evaluations go beyond conventional retrieval benchmarks on top-k recall, offering a more informative indicator of downstream large-model performance and utility in the multimodal domain. We evaluate OpenFlamingo \cite{awadalla2023openflamingo}'s few-shot classification accuracy by using a vision backbone to retrieve a query image's 3 nearest neighbors and prepending the query image with those examples along with their class labels. Following OpenFlamingo's image classification evaluation framework \cite{awadalla2023openflamingo}, we extract the model's logits per object class and determine the model's decision by selecting the class with the largest log probability. See the Appendix for full details on the RAG experimental setup.

As illustrated in Fig.~\ref{fig:rag}, classification accuracy on a wide variety of data domains improves with prompts retrieved by human-aligned models, compared to the original model backbones. Even in out-of-distribution domains such as medical imagery and 2D renders of game scenes, human-aligned models retrieve in-context examples that boost OpenFlamingo classification accuracy. These results suggest that human-aligned models can select more informative examples for in-context learning, thereby boosting the few-shot generalization abilities of a downstream multimodal VLM.

\subsection{Counting}
\label{sec:counting}

\begin{figure*}[ht!]
    \centering
    \includegraphics[width=\linewidth]{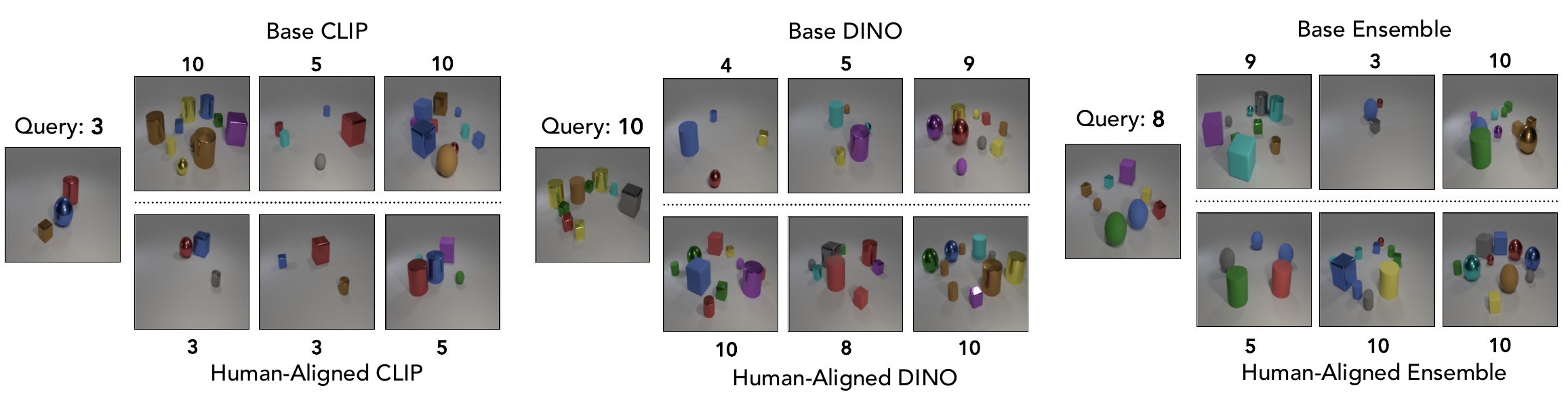}
    \caption{\small \textbf{Visualizations of nearest-neighbor examples retrieved by CLIP, DINO, and Ensemble models as well as their human-aligned versions.} Overall, we see retrieved images with more accurate object counts in CLIP-HA, DINO-HA, and Ensemble-HA across multiple nearest neighbors.}
    \label{fig:clevr}
\end{figure*}

\begin{figure}[ht!]
\RawFloats
    \centering
    \begin{minipage}[c]{0.32\textwidth}
    \centering
        \includegraphics[width=1.0\linewidth]{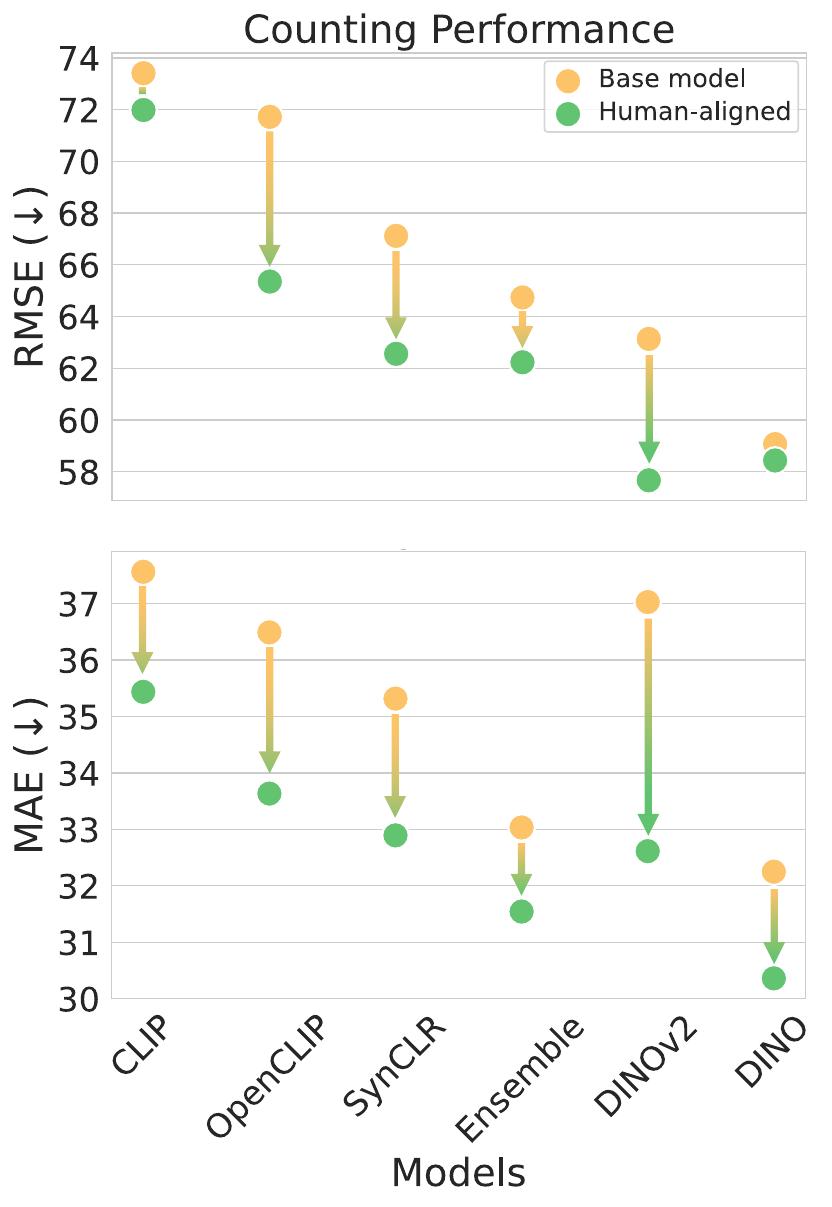}
        \captionof{figure}{\small Performance improvements on Clevr-Count visualized by backbone for RMSE (top) and MAE (bottom), averaged across all datasets. Lower is better.}
        \label{fig:counting}
    \end{minipage} \hfill
    \begin{minipage}[c]{0.63\textwidth}
    \centering
        \resizebox{1.0\linewidth}{!}{
\setlength{\aboverulesep}{0.5pt}
\setlength{\belowrulesep}{0.5pt}
\setlength{\extrarowheight}{.75ex}
\begin{tabular}{l
>{\columncolor{mygray}}c
>{\columncolor{mygray}}ccc
>{\columncolor{mygray}}c
>{\columncolor{mygray}}c}
\toprule
 {} & \multicolumn{2}{c}{\cellcolor{mygray}\bf FSC147} & \multicolumn{2}{c}{\bf CARPK} & \multicolumn{2}{c}{\cellcolor{mygray}\bf Clever-Count} \\ 
 {\bf } & {MAE } & {RMSE } & {MAE } & {RMSE } & {MAE } & {RMSE } \\
 \midrule
  DINO & 44.1 & \bf 118.7 & 51.4 & 56.8 & 1.25 & 1.70 \\
  DINO-HA & \bf{41.3}$^{\dagger}$ & 119.3 & \bf 48.7 & \bf{54.5}$^{\dagger}$ & \bf 1.08 & \bf 1.50 \\
 \midrule
  DINOv2 & 57.5 & 128.3 & 52.4 & 59.5 & 1.18 & 1.60 \\
  DINOv2-HA & \bf 44.9 & \bf{113.6}$^{\dagger}$ & \bf 52.1 & \bf 58.2 & \bf 0.84 & \bf 1.20 \\
 \midrule
  CLIP & 59.1 & 158.4 & 52.5 & 60.3 & 1.09 & 1.52 \\
  CLIP-HA & \bf 53.2 & \bf 156.0 & \bf 52.3 & \bf 58.8 & \bf{0.81} & \bf{1.15}$^{\dagger}$ \\
 \midrule
  OpenCLIP & 54.4 & 153.5 & 54.1 & 60.3 & 0.97 & 1.36\\
  OpenCLIP-HA & \bf 50.2 & \bf 139.1 & \bf 49.9 & \bf 55.8 & \bf 0.80$^{\dagger}$ & \bf 1.15$^{\dagger}$ \\
 \midrule
  SynCLR & 50.6 & 139.6 & 54.3 & 60.3 & 1.05 & 1.45 \\
  SynCLR-HA & \bf 46.4 & \bf 128.1 & \bf 51.3 & \bf 58.2 & \bf 0.98 & \bf 1.37 \\
 \midrule
  Ensemble & 48.4 & 132.4 & 49.6 & 60.3 & 1.10 & 1.51 \\
  Ensemble-HA & \bf 45.4 & \bf 130.3 & \bf{48.4}$^{\dagger}$ & \bf 55.2 & \bf 0.83 & \bf 1.19  \\
 \bottomrule
\end{tabular}
}

        \captionof{table}{\small Error comparisons for base and human-aligned models on standard counting benchmarks. Though FSC147 and CARPK have examples with extreme object counts (tens and hundreds) unseen in the NIGHTS data, human-aligned models still achieve higher performance in each pair. $\dagger$ indicates best score in the column, lower is better.}
        \label{tab:counting}
    \end{minipage}
\end{figure}

A well-documented limitation of large vision backbones is their performance on compositional tasks: in particular, on object counting \cite{Paiss2023TeachingCT}. We investigate how aligning to perceptual judgments affects performance on counting tasks via the FSC147, CARPK, and Clevr-Count (adapted from the original Clevr dataset by \cite{Zhai2019ALS}) benchmarks by computer k-Nearest Neighbors accuracy on frozen vision representations. We report results in Table \ref{tab:counting} and retrieval visualizations for few ($n=3$) and many ($n=8,10$) objects in Fig.~\ref{fig:clevr} and find that, across 6 different models, the human-aligned versions outperform their counterparts in 35 out of 36 cases. See the Appendix for full details on the counting experimental setup.
Given that the perceptual similarity dataset we use for finetuning contains image-level similarity judgments, the consistent improvements that we observe on counting tasks, which requires local object awareness, is somewhat surprising. We hypothesize that the sensitivity of our human-aligned models to object counts may be a byproduct of NIGHTS examples themselves, many of which include image triplets with varying numbers of objects (see Fig.~\ref{fig:dataset} in the Appendix). In terms of human perception, we note that humans consider object count when evaluating image similarity as soon as they develop counting profiency~\cite{MIX1999269}. Thus, it is possible that given the prevalence of triplets with object-count variations in NIGHTS, the human annotations naturally capture this counting aware effect in the global image-level labels and propagate this information to the human-aligned models.

\subsection{Instance retrieval}
\begin{wrapfigure}[16]{r}{0.35\textwidth}
    \vspace{-5mm}
    \resizebox{1.0\linewidth}{!}{
\begin{tabular}{lccc}
\toprule
 {} & \multicolumn{3}{c}{\bf DeepFashion2}  \\
 {\bf } & {Top-1} & {Top-3} & {Top-5} \\
 \midrule
  DINO  & 8.02 & 12.15 & 14.44 \\
  DINO-HA  & \bf 11.69 & \bf 17.55 & \bf 20.84 \\
 \midrule
  DINOv2  & 5.95 & 8.47 & 9.98\\
  DINOv2-HA  & \bf 9.03 & \bf 13.57 & \bf 16.30 \\
 \midrule
  CLIP  & 4.59 & 6.89 & 8.23  \\
  CLIP-HA  & \bf 8.62 & \bf 13.02 & \bf 15.60  \\
 \midrule
  OpenCLIP  & 14.88 & 22.74 & 27.36 \\
  OpenCLIP-HA  & \bf 16.85 & \bf 24.58 & \bf 28.64  \\
 \midrule
  SynCLR  & 4.88 & 7.34 & 9.02\\
  SynCLR-HA  & \bf 8.35 & \bf 12.31 & \bf 14.86\\
 \midrule
  Ensemble  & 13.54 & 20.01 & 23.54\\
  Ensemble-HA  & $\mathbf{23.39}^{\dagger}$ & $\mathbf{32.47}^{\dagger}$ & $\mathbf{37.20}^{\dagger}$ \\
 \bottomrule
\end{tabular}
}

    \captionof{table}{Top-1, -3, and -5 recall scores for instance retrieval on DeepFashion 2. $\dagger$ indicates best score in the column, higher is better.}
    \label{tab:df_table}
\end{wrapfigure}
In this section, we evaluate how aligning models to perceptual judgments affects their performance on the instance retrieval task. This task aims to retrieve images from a gallery containing a shared subject or object with the query image. At test time, retrieval is performed by computing the cosine similarity between the extracted features of the query and gallery images. Succeeding at this task requires that a representation be robust to recognizing instance identities under different lighting, backgrounds, poses, and other such shifts. 

\begin{figure}[H]
\RawFloats
    \centering
    \begin{minipage}{0.41\textwidth}
    \centering
        \includegraphics[width=1.0\linewidth]{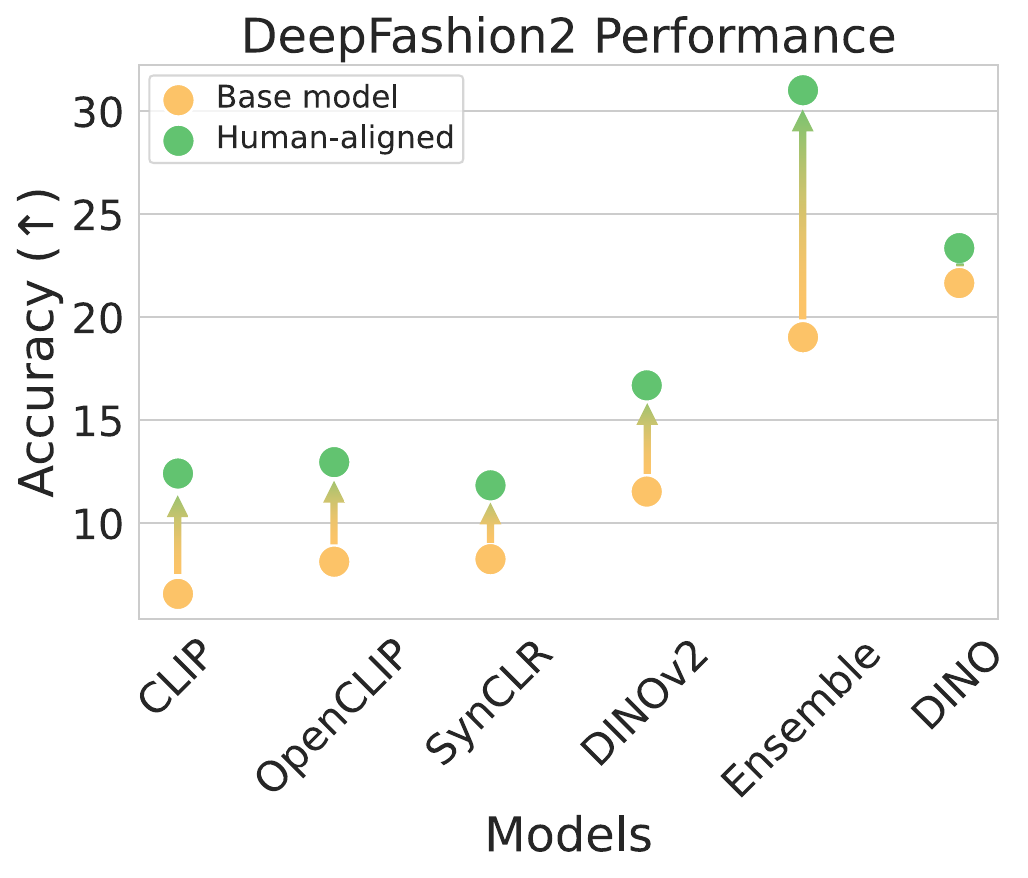}
        \captionof{figure}{\small Performance improvements on the DeepFashion2 instance retrieval, task visualized by backbone and averaged across all $k$ for top-$k$ recall. Higher is better.}
        \label{fig:df_graph}
    \end{minipage}
    \hfill
    \begin{minipage}{0.55\textwidth}
    \centering
        \includegraphics[width=1.0\linewidth]{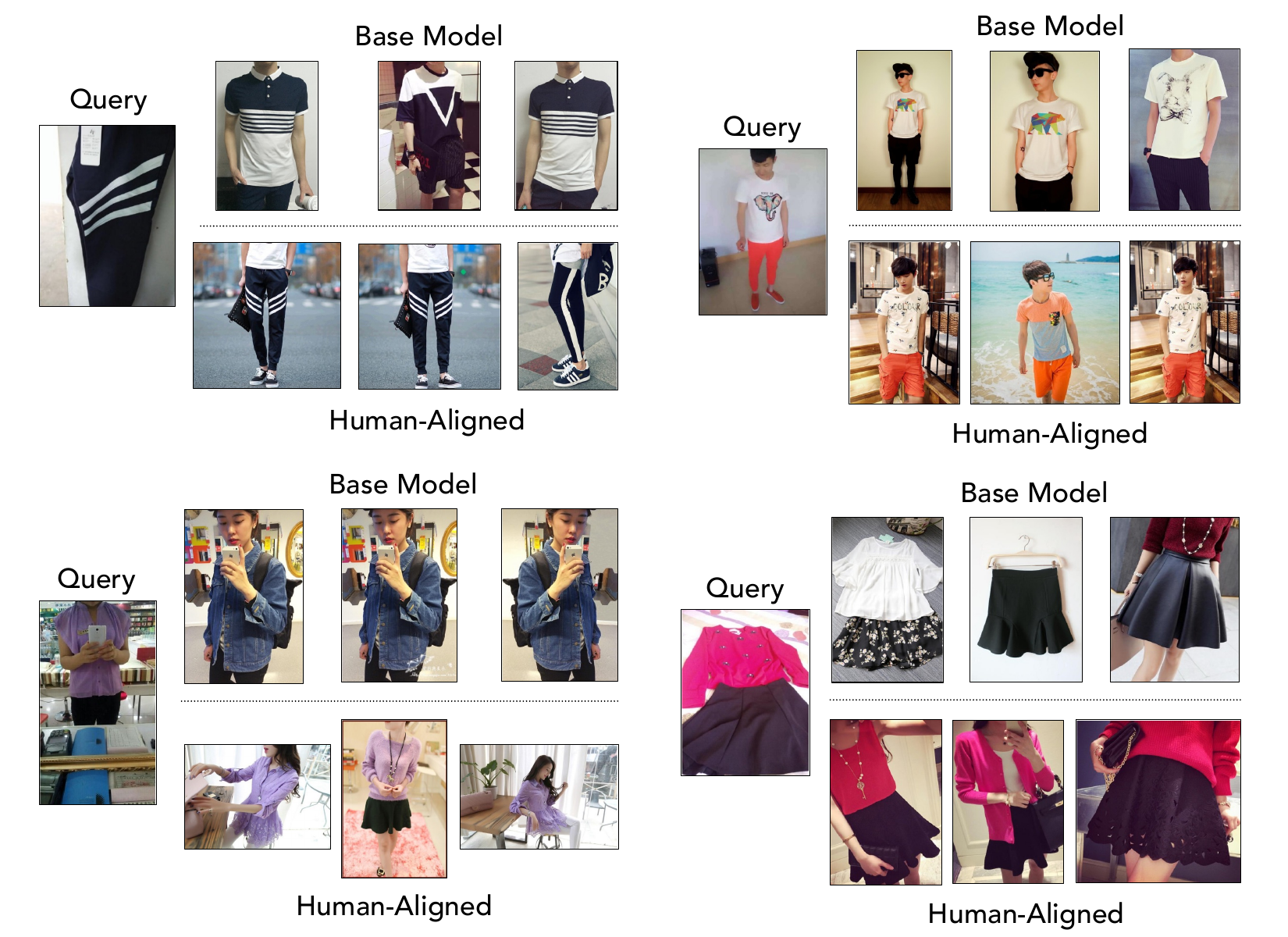}
        \captionof{figure}{\small Examples of top-3 retrievals for a given query image on DeepFashion2. Overall, the human-aligned models return matching clothing items more frequently.}
        \label{fig:df_pics}
    \end{minipage}
\end{figure}

We evaluate all base models and their human-aligned counterparts on the Consumer-to-Shop benchmark of the DeepFashion2 dataset \cite{Ge2019DeepFashion2AV}. The benchmark consists of 10990 consumer "in the wild images" as the query set, and 21438 gallery images with matching clothing items to consumer images. Following the evaluation protocol from \cite{Ge2019DeepFashion2AV}, we report Top-1, 3, and 5 accuracy in Table \ref{tab:df_table}. Human aligned models outperform base models by a significant margin across all metrics and backbones (visualized in Fig.~\ref{fig:df_graph}. In Fig.~\ref{fig:df_pics} we provide qualitative retrieval results. These results agree with prior work showing that training on NIGHTS improves performance in retrieving similar images to queries \cite{dreamsim}. 

\vspace{10mm}

\subsection{What type of human similarity annotation is most beneficial?}
\label{sec:ablation}

\begin{figure*}[ht!]
    \centering
    \includegraphics[width=\linewidth]{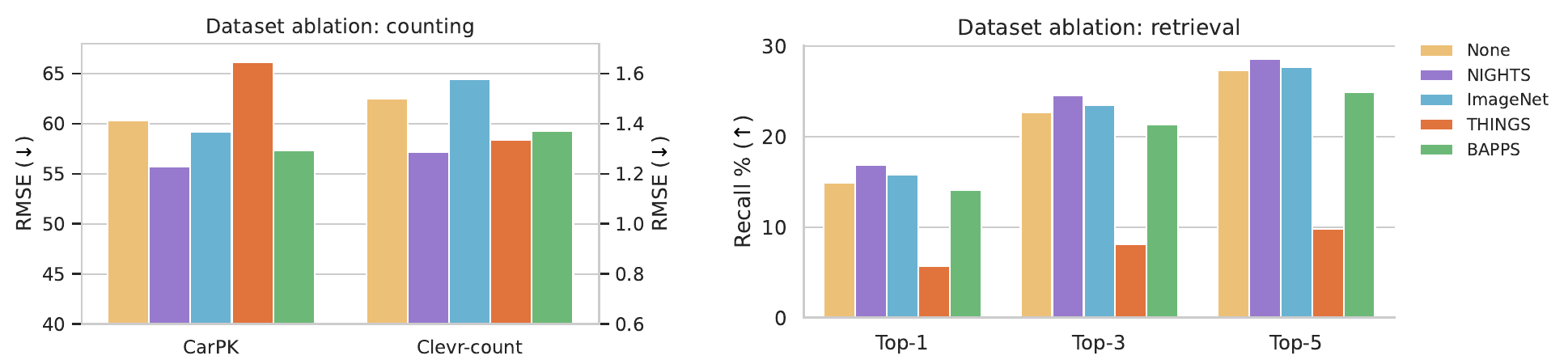}
    \caption{\small Evaluations comparing dataset utility on counting tasks (lower RMSE is better) and DeepFashion2 instance retrieval (higher recall is better). Across each task, tuning on NIGHTS yields the largest improvements while THINGS worsens performance and BAPPS/ImageNet makes minimal changes.}
    \label{fig:ablation}
\end{figure*}

As evidenced by the current widespread use of large vision models \cite{DINO_2021, oquab2023dinov2, clip-models}, trained on massive datasets, model performance largely correlates with data scale.
This raises the question: are our performance gains purely due to training the base models on additional data, or as a result of the perceptual qualities embedded in NIGHTS? To investigate this, we ablate the training dataset - rather than tune on 13,900 NIGHTS triplets, we train on three other image triplet datasets of the same size:
\begin{enumerate}
    \item BAPPS \cite{lpips}: Originally the training set for the LPIPS perceptual similarity metric \cite{lpips}, BAPPS consists of image patch triplets with various low-level distortions applied (e.g. color jitter, gaussian blur, JPEG compression artifacts).
    \item THINGS \cite{things-data}: This dataset contains image triplets with each image encoding a different concept (e.g. a triplet of images categorized as \{airplane, elephant, football\}), labeled by humans tasked to determine which concept is the odd-one-out.
    \item ImageNet \cite{ImageNetChallenge}: To ablate whether perceptual judgments at any level are needed, we construct an image triplet dataset by randomly selecting two images from one category and one image from another, labeling the first image pair as more similar to each other.
\end{enumerate}
For all three datasets, we apply the same training settings as with the original LoRA-tuning on NIGHTS. See Fig.~\ref{fig:ablation} for dataset ablations on object counting and instance retrieval.

Tuning on NIGHTS indeed provides the largest improvements across these tasks, with THINGS worsening performance overall and BAPPS/ImageNet having minimal effect. These trends may appear because BAPPS' photometric distortions are too low-level to impart any perceptual signal onto the backbones, THINGS encodes a higher-level conceptual similarity irrelevant to these mid-level vision tasks, and pre-trained vision models already perform quite well at discriminating ImageNet categories. Indeed, previous work \citep{dreamsim} has found that similarity judgments by perceptual metrics trained on BAPPS correlate better to low-level metrics such as color than to semantic attributes. 

In Section \ref{app-ablation} of the Appendix we find that the performance boost from NIGHTS over other datasets is also consistent across semantic segmentation and depth estimation. Conversely, tuning on NIGHTS fails to improve over other datasets on classification datasets; we further discuss this finding in Sections \ref{discussion} and \ref{sup:vtab_classification}. 

\section{Discussion}
\label{discussion}

Recently, the vision research community has converged on the idea that using human perception to improve machine perception can bolster the transfer of vision representations to downstream tasks~\cite{lpips, NEURIPS2022_harmonizing, muttenthaler2023human, dreamsim, muttenthaler2023improving, muttenthaler2024aligning}. However, it remained unclear which tasks benefit most from alignment with human perceptual judgments; different tasks may demand representations that encode different levels of granularity and semantics. Here, we fine-tune modern backbones, pretrained on different tasks, on human perceptual judgments~\cite{dreamsim} and subsequently investigate which downstream tasks benefit. We develop a better understanding of how perceptual alignment affects performance on important tasks -- e.g., segmentation, depth estimation, RAG -- which may in turn inform future model training and data curation decisions across different applications. 
By evaluating competency at downstream tasks, we quantify what these representations capture and make decodable, enabling a better understanding of the human-aligned feature space. While we probe representations in terms of competency, understanding them in terms of their mechanism is a rich direction for future work.  

We find widespread benefits from perceptual alignment across both image- and patch-level tasks. At the global level, performance consistently improves for retrieval-augmented generation, counting-based and instance-based retrieval. Moreover, propagating the supervision from image-level similarity judgments to ViT patch tokens improves performance on dense prediction tasks (semantic segmentation and depth prediction). Since model performance is closely linked to data scale, we also ablate the choice of perceptual dataset. While fine-tuning on NIGHTS --- a dataset of mid-level perceptual judgments --- leads to improvements across various tasks, fine-tuning on triplets from ImageNet~\cite{ImageNetDatabase}, BAPPS~\cite{lpips}, and THINGS \cite{things-concepts} preserves or deteriorates transfer performance.

Why does fine-tuning on NIGHTS in particular lead to improvements? We hypothesize that the variations found in BAPPS and THINGS are solely high- or low-level, whereas the mid-level distortions in NIGHTS cover salient features that humans use when making inferences about what they see; these characteristics include style, pose, color, and count (see Fig.\ref{fig:dataset}), and largely correlate with the characteristics a model must successfully extract for many computer vision tasks. Previous work \cite{dreamsim} found that models fine-tuned on NIGHTS seem to attend to both low-level and semantic attributes. Aligning a feature space to these concepts may be useful for visual tasks requiring both visual and semantic knowledge, such as retrieval, counting, segmentation, etc. This hypothesis may also explain why tuning on NIGHTS hurts performance on fine-grained tasks, in which perceptually similar images may belong to different categories.

\paragraph{Limitations.} Perceptual alignment does not appear to improve performance for standard image classification tasks such as natural image datasets in the VTAB benchmark~\cite{Zhai2019ALS} (see Tables~\ref{tab:vtab-results}a-b in the Appendix). This is surprising in light of recent findings that demonstrate downstream task improvements in image classification tasks for human-aligned representations~\cite{muttenthaler2023improving}. Although it is hard to pinpoint the exact cause, we hypothesize two reasons: First, perceptual judgments at different levels of abstraction may be helpful for different downstream tasks. While the mid-level perceptual judgments in NIGHTS boost performance for retrieval-based and dense prediction tasks, they may not impart a useful inductive bias for standard image classification tasks; high-level semantic associations could simply be better suited for these kinds of tasks. Alternatively, the visual features that humans use to judge similarity may not be appropriately captured in classification accuracy metrics. Some VTAB datasets have numerical ground truth in which the distance from the correct answer is meaningful; thus, it may be ill-suited for classification, and better suited to continuous evaluations, which we report in sections of the paper. 

Additionally, a key insight from our dataset ablations is that not all human preferences improve performance. Finetuning on perceptual datasets with solely high-level (THINGS) or low-level (BAPPS) variations hurts performance for many downstream tasks (Section ~\ref{sec:ablation} and \ref{app-ablation}). Similarly, \citet{muttenthaler2024aligning} recently discovered that finetuning vision models on THINGS can hurt downstream task transfer. Due to our evaluation of the synthetically-pretrained SynCLR \citep{Tian2023LearningVF}, we attribute the drop in performance to perceptual alignment. Furthermore, by ablating the number of fine-tuning steps in Section \ref{app-ablation}, we show that overfitting to perceptual judgments may also harm downstream performance. 

\paragraph{Societal impacts.} Beyond our findings, there are other possibilities for harm outside the scope of the type of perceptual annotations we have studied:
Human preferences may reflect unwanted biases. A long-standing problem in both language and vision is that biases (e.g. gender, racial) reflected in Internet language/images are inherited by large models, and reflected in their embeddings. One can imagine similar phenomena to happen in visual preferences \citep{NEURIPS2023_73aacd8b}.
Humans may disagree on a preference label, or even disagree with themselves if asked at different points in time. This may lead to noisy data if not filtered carefully, thus harming representations \citep{sucholutsky2023getting, muttenthaler2024aligning}.
Without sufficient demographic diversity in the annotator group, emerging biases may be reflected in the model.  For example, some RLHF-trained language models have been shown to develop a bias towards the opinions of high-income, liberal individuals over their non-RLHF counterparts \citep{pmlr-v202-santurkar23a}.

\section*{Acknowledgements}

We thank Tali Dekel and Richard Zhang for helpful discussions. This work was supported by the Sagol Weizmann-MIT Bridge Program and by a Packard Fellowship and a Sloan Research Fellowship to P.I. and an NSF GRFP Fellowship to S.S.

\newpage
\bibliography{bibliography}

\begin{thebibliography}{62}
\providecommand{\natexlab}[1]{#1}
\providecommand{\url}[1]{\texttt{#1}}
\expandafter\ifx\csname urlstyle\endcsname\relax
  \providecommand{\doi}[1]{doi: #1}\else
  \providecommand{\doi}{doi: \begingroup \urlstyle{rm}\Url}\fi

\bibitem[Alayrac et~al.(2022)Alayrac, Donahue, Luc, Miech, Barr, Hasson, Lenc, Mensch, Millican, Reynolds, Ring, Rutherford, Cabi, Han, Gong, Samangooei, Monteiro, Menick, Borgeaud, Brock, Nematzadeh, Sharifzadeh, Binkowski, Barreira, Vinyals, Zisserman, and Simonyan]{alayrac2022flamingo}
Jean-Baptiste Alayrac, Jeff Donahue, Pauline Luc, Antoine Miech, Iain Barr, Yana Hasson, Karel Lenc, Arthur Mensch, Katie Millican, Malcolm Reynolds, Roman Ring, Eliza Rutherford, Serkan Cabi, Tengda Han, Zhitao Gong, Sina Samangooei, Marianne Monteiro, Jacob Menick, Sebastian Borgeaud, Andrew Brock, Aida Nematzadeh, Sahand Sharifzadeh, Mikolaj Binkowski, Ricardo Barreira, Oriol Vinyals, Andrew Zisserman, and Karen Simonyan.
\newblock Flamingo: a visual language model for few-shot learning, 2022.

\bibitem[Awadalla et~al.(2023)Awadalla, Gao, Gardner, Hessel, Hanafy, Zhu, Marathe, Bitton, Gadre, Sagawa, Jitsev, Kornblith, Koh, Ilharco, Wortsman, and Schmidt]{awadalla2023openflamingo}
Anas Awadalla, Irena Gao, Josh Gardner, Jack Hessel, Yusuf Hanafy, Wanrong Zhu, Kalyani Marathe, Yonatan Bitton, Samir Gadre, Shiori Sagawa, Jenia Jitsev, Simon Kornblith, Pang~Wei Koh, Gabriel Ilharco, Mitchell Wortsman, and Ludwig Schmidt.
\newblock Openflamingo: An open-source framework for training large autoregressive vision-language models, 2023.

\bibitem[Bai et~al.(2022)Bai, Jones, Ndousse, Askell, Chen, DasSarma, Drain, Fort, Ganguli, Henighan, et~al.]{bai2022training}
Yuntao Bai, Andy Jones, Kamal Ndousse, Amanda Askell, Anna Chen, Nova DasSarma, Dawn Drain, Stanislav Fort, Deep Ganguli, Tom Henighan, et~al.
\newblock Training a helpful and harmless assistant with reinforcement learning from human feedback.
\newblock \emph{arXiv preprint arXiv:2204.05862}, 2022.

\bibitem[Caron et~al.(2021)Caron, Touvron, Misra, J\'egou, Mairal, Bojanowski, and Joulin]{DINO_2021}
Mathilde Caron, Hugo Touvron, Ishan Misra, Herv\'e J\'egou, Julien Mairal, Piotr Bojanowski, and Armand Joulin.
\newblock Emerging properties in self-supervised vision transformers.
\newblock In \emph{Proceedings of the IEEE/CVF International Conference on Computer Vision (ICCV)}, pp.\  9650--9660, October 2021.

\bibitem[Chechik et~al.(2010)Chechik, Sharma, Shalit, and Bengio]{chechik2010large}
Gal Chechik, Varun Sharma, Uri Shalit, and Samy Bengio.
\newblock Large scale online learning of image similarity through ranking.
\newblock \emph{Journal of Machine Learning Research}, 11\penalty0 (3), 2010.

\bibitem[Chen et~al.(2020{\natexlab{a}})Chen, Kornblith, Norouzi, and Hinton]{simclr}
Ting Chen, Simon Kornblith, Mohammad Norouzi, and Geoffrey Hinton.
\newblock A simple framework for contrastive learning of visual representations.
\newblock In Hal~Daumé III and Aarti Singh (eds.), \emph{Proceedings of the 37th International Conference on Machine Learning}, volume 119 of \emph{Proceedings of Machine Learning Research}, pp.\  1597--1607. PMLR, 13--18 Jul 2020{\natexlab{a}}.

\bibitem[Chen et~al.(2020{\natexlab{b}})Chen, Kornblith, Swersky, Norouzi, and Hinton]{simclrv2}
Ting Chen, Simon Kornblith, Kevin Swersky, Mohammad Norouzi, and Geoffrey~E Hinton.
\newblock Big self-supervised models are strong semi-supervised learners.
\newblock In H.~Larochelle, M.~Ranzato, R.~Hadsell, M.F. Balcan, and H.~Lin (eds.), \emph{Advances in Neural Information Processing Systems}, volume~33, pp.\  22243--22255. Curran Associates, Inc., 2020{\natexlab{b}}.

\bibitem[Cherti et~al.(2023)Cherti, Beaumont, Wightman, Wortsman, Ilharco, Gordon, Schuhmann, Schmidt, and Jitsev]{openclip}
Mehdi Cherti, Romain Beaumont, Ross Wightman, Mitchell Wortsman, Gabriel Ilharco, Cade Gordon, Christoph Schuhmann, Ludwig Schmidt, and Jenia Jitsev.
\newblock Reproducible scaling laws for contrastive language-image learning.
\newblock In \emph{Proceedings of the IEEE/CVF Conference on Computer Vision and Pattern Recognition}, pp.\  2818--2829, 2023.

\bibitem[Christiano et~al.(2017)Christiano, Leike, Brown, Martic, Legg, and Amodei]{rlhp2017}
Paul~F Christiano, Jan Leike, Tom Brown, Miljan Martic, Shane Legg, and Dario Amodei.
\newblock Deep reinforcement learning from human preferences.
\newblock In I.~Guyon, U.~Von Luxburg, S.~Bengio, H.~Wallach, R.~Fergus, S.~Vishwanathan, and R.~Garnett (eds.), \emph{Advances in Neural Information Processing Systems}, volume~30. Curran Associates, Inc., 2017.

\bibitem[Deng et~al.(2009)Deng, Dong, Socher, Li, Li, and Fei-Fei]{ImageNetDatabase}
Jia Deng, Wei Dong, Richard Socher, Li-Jia Li, Kai Li, and Li~Fei-Fei.
\newblock {I}mage{N}et: A large-scale hierarchical image database.
\newblock In \emph{2009 IEEE Conference on Computer Vision and Pattern Recognition}, pp.\  248--255, 2009.
\newblock \doi{10.1109/CVPR.2009.5206848}.

\bibitem[Ding et~al.(2023)Ding, Zhang, Clune, Spector, and Lehman]{ding2023quality}
Li~Ding, Jenny Zhang, Jeff Clune, Lee Spector, and Joel Lehman.
\newblock Quality diversity through human feedback.
\newblock \emph{arXiv preprint arXiv:2310.12103}, 2023.

\bibitem[Donahue et~al.(2014)Donahue, Jia, Vinyals, Hoffman, Zhang, Tzeng, and Darrell]{donahue2014decaf}
Jeff Donahue, Yangqing Jia, Oriol Vinyals, Judy Hoffman, Ning Zhang, Eric Tzeng, and Trevor Darrell.
\newblock Decaf: A deep convolutional activation feature for generic visual recognition.
\newblock In \emph{International conference on machine learning}, pp.\  647--655. PMLR, 2014.

\bibitem[Dosovitskiy et~al.(2021)Dosovitskiy, Beyer, Kolesnikov, Weissenborn, Zhai, Unterthiner, Dehghani, Minderer, Heigold, Gelly, Uszkoreit, and Houlsby]{vit}
Alexey Dosovitskiy, Lucas Beyer, Alexander Kolesnikov, Dirk Weissenborn, Xiaohua Zhai, Thomas Unterthiner, Mostafa Dehghani, Matthias Minderer, Georg Heigold, Sylvain Gelly, Jakob Uszkoreit, and Neil Houlsby.
\newblock An image is worth 16x16 words: Transformers for image recognition at scale.
\newblock In \emph{International Conference on Learning Representations}, 2021.

\bibitem[Dwibedi et~al.(2021)Dwibedi, Aytar, Tompson, Sermanet, and Zisserman]{dwibedi2021little}
Debidatta Dwibedi, Yusuf Aytar, Jonathan Tompson, Pierre Sermanet, and Andrew Zisserman.
\newblock With a little help from my friends: Nearest-neighbor contrastive learning of visual representations.
\newblock In \emph{Proceedings of the IEEE/CVF International Conference on Computer Vision}, pp.\  9588--9597, 2021.

\bibitem[Eigen et~al.(2014)Eigen, Puhrsch, and Fergus]{eigen2014depth}
David Eigen, Christian Puhrsch, and Rob Fergus.
\newblock Depth map prediction from a single image using a multi-scale deep network, 2014.

\bibitem[Fan et~al.(2023)Fan, Watkins, Du, Liu, Ryu, Boutilier, Abbeel, Ghavamzadeh, Lee, and Lee]{fan2023dpok}
Ying Fan, Olivia Watkins, Yuqing Du, Hao Liu, Moonkyung Ryu, Craig Boutilier, Pieter Abbeel, Mohammad Ghavamzadeh, Kangwook Lee, and Kimin Lee.
\newblock Dpok: Reinforcement learning for fine-tuning text-to-image diffusion models, 2023.

\bibitem[FEL et~al.(2022)FEL, Rodriguez~Rodriguez, Linsley, and Serre]{NEURIPS2022_harmonizing}
Thomas FEL, Ivan~F Rodriguez~Rodriguez, Drew Linsley, and Thomas Serre.
\newblock Harmonizing the object recognition strategies of deep neural networks with humans.
\newblock In S.~Koyejo, S.~Mohamed, A.~Agarwal, D.~Belgrave, K.~Cho, and A.~Oh (eds.), \emph{Advances in Neural Information Processing Systems}, volume~35, pp.\  9432--9446. Curran Associates, Inc., 2022.

\bibitem[Fu et~al.(2023)Fu, Tamir, Sundaram, Chai, Zhang, Dekel, and Isola]{dreamsim}
Stephanie Fu, Netanel Tamir, Shobhita Sundaram, Lucy Chai, Richard Zhang, Tali Dekel, and Phillip Isola.
\newblock Dreamsim: Learning new dimensions of human visual similarity using synthetic data.
\newblock In A.~Oh, T.~Neumann, A.~Globerson, K.~Saenko, M.~Hardt, and S.~Levine (eds.), \emph{Advances in Neural Information Processing Systems}, volume~36, pp.\  50742--50768. Curran Associates, Inc., 2023.

\bibitem[Gatys et~al.(2015{\natexlab{a}})Gatys, Ecker, and Bethge]{gatys2015texture}
Leon Gatys, Alexander~S Ecker, and Matthias Bethge.
\newblock Texture synthesis using convolutional neural networks.
\newblock \emph{Advances in neural information processing systems}, 28, 2015{\natexlab{a}}.

\bibitem[Gatys et~al.(2015{\natexlab{b}})Gatys, Ecker, and Bethge]{gatys2015neural}
Leon~A Gatys, Alexander~S Ecker, and Matthias Bethge.
\newblock A neural algorithm of artistic style.
\newblock \emph{arXiv preprint arXiv:1508.06576}, 2015{\natexlab{b}}.

\bibitem[Ge et~al.(2019)Ge, Zhang, Wu, Wang, Tang, and Luo]{Ge2019DeepFashion2AV}
Yuying Ge, Ruimao Zhang, Lingyun Wu, Xiaogang Wang, Xiaoou Tang, and Ping Luo.
\newblock Deepfashion2: A versatile benchmark for detection, pose estimation, segmentation and re-identification of clothing images.
\newblock \emph{2019 IEEE/CVF Conference on Computer Vision and Pattern Recognition (CVPR)}, pp.\  5332--5340, 2019.

\bibitem[He et~al.(2020)He, Fan, Wu, Xie, and Girshick]{moco}
Kaiming He, Haoqi Fan, Yuxin Wu, Saining Xie, and Ross Girshick.
\newblock {M}omentum {C}ontrast for {U}nsupervised {V}isual {R}epresentation {L}earning.
\newblock In \emph{Proceedings of the IEEE/CVF Conference on Computer Vision and Pattern Recognition (CVPR)}, June 2020.

\bibitem[Hebart et~al.(2020)Hebart, Zheng, Pereira, and Baker]{things-concepts}
Martin~N. Hebart, Charles~Y. Zheng, Francisco Pereira, and Chris~I. Baker.
\newblock Revealing the multidimensional mental representations of natural objects underlying human similarity judgements.
\newblock \emph{Nature Human Behaviour}, 4\penalty0 (11):\penalty0 1173--1185, October 2020.
\newblock \doi{10.1038/s41562-020-00951-3}.

\bibitem[Hebart et~al.(2023)Hebart, Contier, Teichmann, Rockter, Zheng, Kidder, Corriveau, Vaziri-Pashkam, and Baker]{things-data}
Martin~N Hebart, Oliver Contier, Lina Teichmann, Adam~H Rockter, Charles~Y Zheng, Alexis Kidder, Anna Corriveau, Maryam Vaziri-Pashkam, and Chris~I Baker.
\newblock Things-data, a multimodal collection of large-scale datasets for investigating object representations in human brain and behavior.
\newblock \emph{eLife}, 12:\penalty0 e82580, feb 2023.
\newblock ISSN 2050-084X.
\newblock \doi{10.7554/eLife.82580}.

\bibitem[Jahanian et~al.(2021)Jahanian, Puig, Tian, and Isola]{jahanian2021generative}
Ali Jahanian, Xavier Puig, Yonglong Tian, and Phillip Isola.
\newblock Generative models as a data source for multiview representation learning.
\newblock \emph{arXiv preprint arXiv:2106.05258}, 2021.

\bibitem[Jia et~al.(2021)Jia, Yang, Xia, Chen, Parekh, Pham, Le, Sung, Li, and Duerig]{align}
Chao Jia, Yinfei Yang, Ye~Xia, Yi-Ting Chen, Zarana Parekh, Hieu Pham, Quoc Le, Yun-Hsuan Sung, Zhen Li, and Tom Duerig.
\newblock Scaling up visual and vision-language representation learning with noisy text supervision.
\newblock In Marina Meila and Tong Zhang (eds.), \emph{Proceedings of the 38th International Conference on Machine Learning}, volume 139 of \emph{Proceedings of Machine Learning Research}, pp.\  4904--4916. PMLR, 18--24 Jul 2021.

\bibitem[Johnson et~al.(2016)Johnson, Alahi, and Fei-Fei]{johnson2016perceptual}
Justin Johnson, Alexandre Alahi, and Li~Fei-Fei.
\newblock Perceptual losses for real-time style transfer and super-resolution.
\newblock In \emph{Computer Vision--ECCV 2016: 14th European Conference, Amsterdam, The Netherlands, October 11-14, 2016, Proceedings, Part II 14}, pp.\  694--711. Springer, 2016.

\bibitem[Khosla et~al.(2020)Khosla, Teterwak, Wang, Sarna, Tian, Isola, Maschinot, Liu, and Krishnan]{khosla2020supervised}
Prannay Khosla, Piotr Teterwak, Chen Wang, Aaron Sarna, Yonglong Tian, Phillip Isola, Aaron Maschinot, Ce~Liu, and Dilip Krishnan.
\newblock Supervised contrastive learning.
\newblock \emph{Advances in neural information processing systems}, 33:\penalty0 18661--18673, 2020.

\bibitem[Kirstain et~al.(2023{\natexlab{a}})Kirstain, Polyak, Singer, Matiana, Penna, and Levy]{NEURIPS2023_73aacd8b}
Yuval Kirstain, Adam Polyak, Uriel Singer, Shahbuland Matiana, Joe Penna, and Omer Levy.
\newblock Pick-a-pic: An open dataset of user preferences for text-to-image generation.
\newblock In A.~Oh, T.~Naumann, A.~Globerson, K.~Saenko, M.~Hardt, and S.~Levine (eds.), \emph{Advances in Neural Information Processing Systems}, volume~36, pp.\  36652--36663. Curran Associates, Inc., 2023{\natexlab{a}}.
\newblock URL \url{https://proceedings.neurips.cc/paper_files/paper/2023/file/73aacd8b3b05b4b503d58310b523553c-Paper-Conference.pdf}.

\bibitem[Kirstain et~al.(2023{\natexlab{b}})Kirstain, Polyak, Singer, Matiana, Penna, and Levy]{kirstain2023pickapic}
Yuval Kirstain, Adam Polyak, Uriel Singer, Shahbuland Matiana, Joe Penna, and Omer Levy.
\newblock Pick-a-pic: An open dataset of user preferences for text-to-image generation, 2023{\natexlab{b}}.

\bibitem[Kobayashi et~al.(2022)Kobayashi, Matsumoto, and Sitzmann]{kobayashi2022decomposing}
Sosuke Kobayashi, Eiichi Matsumoto, and Vincent Sitzmann.
\newblock Decomposing nerf for editing via feature field distillation.
\newblock \emph{Advances in Neural Information Processing Systems}, 35:\penalty0 23311--23330, 2022.

\bibitem[Laurençon et~al.(2023)Laurençon, Saulnier, Tronchon, Bekman, Singh, Lozhkov, Wang, Karamcheti, Rush, Kiela, Cord, and Sanh]{laurencon2023obelics}
Hugo Laurençon, Lucile Saulnier, Léo Tronchon, Stas Bekman, Amanpreet Singh, Anton Lozhkov, Thomas Wang, Siddharth Karamcheti, Alexander~M. Rush, Douwe Kiela, Matthieu Cord, and Victor Sanh.
\newblock Obelics: An open web-scale filtered dataset of interleaved image-text documents, 2023.

\bibitem[Laurençon et~al.(2024)Laurençon, Tronchon, Cord, and Sanh]{laurençon2024matters}
Hugo Laurençon, Léo Tronchon, Matthieu Cord, and Victor Sanh.
\newblock What matters when building vision-language models?, 2024.

\bibitem[Lewis et~al.(2021)Lewis, Perez, Piktus, Petroni, Karpukhin, Goyal, Küttler, Lewis, tau Yih, Rocktäschel, Riedel, and Kiela]{lewis2021retrievalaugmented}
Patrick Lewis, Ethan Perez, Aleksandra Piktus, Fabio Petroni, Vladimir Karpukhin, Naman Goyal, Heinrich Küttler, Mike Lewis, Wen tau Yih, Tim Rocktäschel, Sebastian Riedel, and Douwe Kiela.
\newblock Retrieval-augmented generation for knowledge-intensive nlp tasks, 2021.

\bibitem[Li et~al.(2019)Li, Luo, Lu, Xiang, and Wang]{li2019large}
Aoxue Li, Tiange Luo, Zhiwu Lu, Tao Xiang, and Liwei Wang.
\newblock Large-scale few-shot learning: Knowledge transfer with class hierarchy.
\newblock In \emph{Proceedings of the ieee/cvf conference on computer vision and pattern recognition}, pp.\  7212--7220, 2019.

\bibitem[Li \& Snavely(2018)Li and Snavely]{MegaDepthLi18}
Zhengqi Li and Noah Snavely.
\newblock Megadepth: Learning single-view depth prediction from internet photos.
\newblock In \emph{Computer Vision and Pattern Recognition (CVPR)}, 2018.

\bibitem[Li et~al.(2022)Li, Wang, Liu, and Jiang]{li2022binsformer}
Zhenyu Li, Xuyang Wang, Xianming Liu, and Junjun Jiang.
\newblock Binsformer: Revisiting adaptive bins for monocular depth estimation, 2022.

\bibitem[Liang et~al.(2023)Liang, Laidlaw, Meyerowitz, Sridhar, and Tompkin]{liang2023semantic}
Yiqing Liang, Eliot Laidlaw, Alexander Meyerowitz, Srinath Sridhar, and James Tompkin.
\newblock Semantic attention flow fields for monocular dynamic scene decomposition.
\newblock In \emph{Proceedings of the IEEE/CVF International Conference on Computer Vision}, pp.\  21797--21806, 2023.

\bibitem[Luo et~al.(2024)Luo, Dunlap, Park, Holynski, and Darrell]{luo2024diffusion}
Grace Luo, Lisa Dunlap, Dong~Huk Park, Aleksander Holynski, and Trevor Darrell.
\newblock Diffusion hyperfeatures: Searching through time and space for semantic correspondence.
\newblock \emph{Advances in Neural Information Processing Systems}, 36, 2024.

\bibitem[Mix(1999)]{MIX1999269}
Kelly~S. Mix.
\newblock Similarity and numerical equivalence: Appearances count.
\newblock \emph{Cognitive Development}, 14\penalty0 (2):\penalty0 269--297, 1999.
\newblock ISSN 0885-2014.
\newblock \doi{https://doi.org/10.1016/S0885-2014(99)00005-2}.

\bibitem[Muttenthaler et~al.(2023{\natexlab{a}})Muttenthaler, Dippel, Linhardt, Vandermeulen, and Kornblith]{muttenthaler2023human}
Lukas Muttenthaler, Jonas Dippel, Lorenz Linhardt, Robert~A Vandermeulen, and Simon Kornblith.
\newblock Human alignment of neural network representations.
\newblock In \emph{11th International Conference on Learning Representations, {ICLR} 2023, Kigali, Rwanda, Mai 01-05, 2023}. OpenReview.net, 2023{\natexlab{a}}.

\bibitem[Muttenthaler et~al.(2023{\natexlab{b}})Muttenthaler, Linhardt, Dippel, Vandermeulen, Hermann, Lampinen, and Kornblith]{muttenthaler2023improving}
Lukas Muttenthaler, Lorenz Linhardt, Jonas Dippel, Robert~A Vandermeulen, Katherine Hermann, Andrew Lampinen, and Simon Kornblith.
\newblock Improving neural network representations using human similarity judgments.
\newblock In A.~Oh, T.~Neumann, A.~Globerson, K.~Saenko, M.~Hardt, and S.~Levine (eds.), \emph{Advances in Neural Information Processing Systems}, volume~36, pp.\  50978--51007. Curran Associates, Inc., 2023{\natexlab{b}}.

\bibitem[Muttenthaler et~al.(2024)Muttenthaler, Greff, Born, Spitzer, Kornblith, Mozer, M{\"u}ller, Unterthiner, and Lampinen]{muttenthaler2024aligning}
Lukas Muttenthaler, Klaus Greff, Frieda Born, Bernhard Spitzer, Simon Kornblith, Michael~C Mozer, Klaus-Robert M{\"u}ller, Thomas Unterthiner, and Andrew~K Lampinen.
\newblock Aligning machine and human visual representations across abstraction levels.
\newblock \emph{arXiv preprint arXiv:2409.06509}, 2024.

\bibitem[Oquab et~al.(2023)Oquab, Darcet, Moutakanni, Vo, Szafraniec, Khalidov, Fernandez, Haziza, Massa, El-Nouby, Howes, Huang, Xu, Sharma, Li, Galuba, Rabbat, Assran, Ballas, Synnaeve, Misra, Jegou, Mairal, Labatut, Joulin, and Bojanowski]{oquab2023dinov2}
Maxime Oquab, Timothée Darcet, Theo Moutakanni, Huy~V. Vo, Marc Szafraniec, Vasil Khalidov, Pierre Fernandez, Daniel Haziza, Francisco Massa, Alaaeldin El-Nouby, Russell Howes, Po-Yao Huang, Hu~Xu, Vasu Sharma, Shang-Wen Li, Wojciech Galuba, Mike Rabbat, Mido Assran, Nicolas Ballas, Gabriel Synnaeve, Ishan Misra, Herve Jegou, Julien Mairal, Patrick Labatut, Armand Joulin, and Piotr Bojanowski.
\newblock Dinov2: Learning robust visual features without supervision, 2023.

\bibitem[Paiss et~al.(2023)Paiss, Ephrat, Tov, Zada, Mosseri, Irani, and Dekel]{Paiss2023TeachingCT}
Roni Paiss, Ariel Ephrat, Omer Tov, Shiran Zada, Inbar Mosseri, Michal Irani, and Tali Dekel.
\newblock Teaching clip to count to ten.
\newblock \emph{2023 IEEE/CVF International Conference on Computer Vision (ICCV)}, pp.\  3147--3157, 2023.

\bibitem[Radford et~al.(2021)Radford, Kim, Hallacy, Ramesh, Goh, Agarwal, Sastry, Askell, Mishkin, Clark, Krueger, and Sutskever]{clip-models}
Alec Radford, Jong~Wook Kim, Chris Hallacy, Aditya Ramesh, Gabriel Goh, Sandhini Agarwal, Girish Sastry, Amanda Askell, Pamela Mishkin, Jack Clark, Gretchen Krueger, and Ilya Sutskever.
\newblock Learning transferable visual models from natural language supervision.
\newblock In Marina Meila and Tong Zhang (eds.), \emph{Proceedings of the 38th International Conference on Machine Learning}, volume 139 of \emph{Proceedings of Machine Learning Research}, pp.\  8748--8763. PMLR, 18--24 Jul 2021.

\bibitem[Robinson et~al.(2020)Robinson, Chuang, Sra, and Jegelka]{robinson2020contrastive}
Joshua Robinson, Ching-Yao Chuang, Suvrit Sra, and Stefanie Jegelka.
\newblock Contrastive learning with hard negative samples.
\newblock \emph{arXiv preprint arXiv:2010.04592}, 2020.

\bibitem[Russakovsky et~al.(2015)Russakovsky, Deng, Su, Krause, Satheesh, Ma, Huang, Karpathy, Khosla, Bernstein, Berg, and Li]{ImageNetChallenge}
Olga Russakovsky, Jia Deng, Hao Su, Jonathan Krause, Sanjeev Satheesh, Sean Ma, Zhiheng Huang, Andrej Karpathy, Aditya Khosla, Michael~S. Bernstein, Alexander~C. Berg, and Fei{-}Fei Li.
\newblock {I}mage{N}et large scale visual recognition challenge.
\newblock \emph{Int. J. Comput. Vis.}, 115\penalty0 (3):\penalty0 211--252, 2015.
\newblock \doi{10.1007/s11263-015-0816-y}.

\bibitem[Santurkar et~al.(2023)Santurkar, Durmus, Ladhak, Lee, Liang, and Hashimoto]{pmlr-v202-santurkar23a}
Shibani Santurkar, Esin Durmus, Faisal Ladhak, Cinoo Lee, Percy Liang, and Tatsunori Hashimoto.
\newblock Whose opinions do language models reflect?
\newblock In Andreas Krause, Emma Brunskill, Kyunghyun Cho, Barbara Engelhardt, Sivan Sabato, and Jonathan Scarlett (eds.), \emph{Proceedings of the 40th International Conference on Machine Learning}, volume 202 of \emph{Proceedings of Machine Learning Research}, pp.\  29971--30004. PMLR, 23--29 Jul 2023.
\newblock URL \url{https://proceedings.mlr.press/v202/santurkar23a.html}.

\bibitem[Sharif~Razavian et~al.(2014)Sharif~Razavian, Azizpour, Sullivan, and Carlsson]{sharif2014cnn}
Ali Sharif~Razavian, Hossein Azizpour, Josephine Sullivan, and Stefan Carlsson.
\newblock Cnn features off-the-shelf: an astounding baseline for recognition.
\newblock In \emph{Proceedings of the IEEE conference on computer vision and pattern recognition workshops}, pp.\  806--813, 2014.

\bibitem[Sharma et~al.(2023)Sharma, Philip, Gharbi, Freeman, Durand, and Deschaintre]{sharma2023materialistic}
Prafull Sharma, Julien Philip, Micha{\"e}l Gharbi, Bill Freeman, Fredo Durand, and Valentin Deschaintre.
\newblock Materialistic: Selecting similar materials in images.
\newblock \emph{ACM Transactions on Graphics (TOG)}, 42\penalty0 (4):\penalty0 1--14, 2023.

\bibitem[Shen et~al.(2023)Shen, Yang, Yu, Wong, Kaelbling, and Isola]{shen2023distilled}
William Shen, Ge~Yang, Alan Yu, Jansen Wong, Leslie~Pack Kaelbling, and Phillip Isola.
\newblock Distilled feature fields enable few-shot language-guided manipulation.
\newblock \emph{arXiv preprint arXiv:2308.07931}, 2023.

\bibitem[Sohn et~al.(2023)Sohn, Ruiz, Lee, Chin, Blok, Chang, Barber, Jiang, Entis, Li, et~al.]{sohn2023styledrop}
Kihyuk Sohn, Nataniel Ruiz, Kimin Lee, Daniel~Castro Chin, Irina Blok, Huiwen Chang, Jarred Barber, Lu~Jiang, Glenn Entis, Yuanzhen Li, et~al.
\newblock Styledrop: Text-to-image generation in any style.
\newblock \emph{arXiv preprint arXiv:2306.00983}, 2023.

\bibitem[Sucholutsky et~al.(2023)Sucholutsky, Muttenthaler, Weller, Peng, Bobu, Kim, Love, Grant, Achterberg, Tenenbaum, et~al.]{sucholutsky2023getting}
Ilia Sucholutsky, Lukas Muttenthaler, Adrian Weller, Andi Peng, Andreea Bobu, Been Kim, Bradley~C Love, Erin Grant, Jascha Achterberg, Joshua~B Tenenbaum, et~al.
\newblock Getting aligned on representational alignment.
\newblock \emph{arXiv preprint arXiv:2310.13018}, 2023.

\bibitem[Tian et~al.(2023)Tian, Fan, Chen, Katabi, Krishnan, and Isola]{Tian2023LearningVF}
Yonglong Tian, Lijie Fan, Kaifeng Chen, Dina Katabi, Dilip Krishnan, and Phillip Isola.
\newblock Learning vision from models rivals learning vision from data.
\newblock \emph{ArXiv}, abs/2312.17742, 2023.

\bibitem[Tong et~al.(2024)Tong, Liu, Zhai, Ma, LeCun, and Xie]{Tong_2024_CVPR}
Shengbang Tong, Zhuang Liu, Yuexiang Zhai, Yi~Ma, Yann LeCun, and Saining Xie.
\newblock Eyes wide shut? exploring the visual shortcomings of multimodal llms.
\newblock In \emph{Proceedings of the IEEE/CVF Conference on Computer Vision and Pattern Recognition (CVPR)}, pp.\  9568--9578, June 2024.

\bibitem[Tumanyan et~al.(2024)Tumanyan, Singer, Bagon, and Dekel]{tumanyan2024dino}
Narek Tumanyan, Assaf Singer, Shai Bagon, and Tali Dekel.
\newblock Dino-tracker: Taming dino for self-supervised point tracking in a single video.
\newblock \emph{arXiv preprint arXiv:2403.14548}, 2024.

\bibitem[Wu et~al.(2023)Wu, Sun, Zhu, Zhao, and Li]{hps_2023}
Xiaoshi Wu, Keqiang Sun, Feng Zhu, Rui Zhao, and Hongsheng Li.
\newblock Human preference score: Better aligning text-to-image models with human preference.
\newblock In \emph{{IEEE/CVF} International Conference on Computer Vision, {ICCV} 2023, Paris, France, October 1-6, 2023}, pp.\  2096--2105. {IEEE}, 2023.
\newblock \doi{10.1109/ICCV51070.2023.00200}.

\bibitem[Zbontar et~al.(2021)Zbontar, Jing, Misra, LeCun, and Deny]{barlowtwins}
Jure Zbontar, Li~Jing, Ishan Misra, Yann LeCun, and Stephane Deny.
\newblock Barlow twins: Self-supervised learning via redundancy reduction.
\newblock In Marina Meila and Tong Zhang (eds.), \emph{Proceedings of the 38th International Conference on Machine Learning}, volume 139 of \emph{Proceedings of Machine Learning Research}, pp.\  12310--12320. PMLR, 18--24 Jul 2021.

\bibitem[Zhai et~al.(2019)Zhai, Puigcerver, Kolesnikov, Ruyssen, Riquelme, Lucic, Djolonga, Pinto, Neumann, Dosovitskiy, Beyer, Bachem, Tschannen, Michalski, Bousquet, Gelly, and Houlsby]{Zhai2019ALS}
Xiaohua Zhai, Joan Puigcerver, Alexander Kolesnikov, Pierre Ruyssen, Carlos Riquelme, Mario Lucic, Josip Djolonga, Andr{\'e}~Susano Pinto, Maxim Neumann, Alexey Dosovitskiy, Lucas Beyer, Olivier Bachem, Michael Tschannen, Marcin Michalski, Olivier Bousquet, Sylvain Gelly, and Neil Houlsby.
\newblock A large-scale study of representation learning with the visual task adaptation benchmark.
\newblock \emph{arXiv: Computer Vision and Pattern Recognition}, 2019.

\bibitem[Zhang et~al.(2018)Zhang, Isola, Efros, Shechtman, and Wang]{lpips}
Richard Zhang, Phillip Isola, Alexei~A. Efros, Eli Shechtman, and Oliver Wang.
\newblock The unreasonable effectiveness of deep features as a perceptual metric.
\newblock In \emph{Proceedings of the IEEE Conference on Computer Vision and Pattern Recognition (CVPR)}, June 2018.

\bibitem[Zhang et~al.(2023)Zhang, Yang, Feng, Qin, Chen, Yu, Chen, Wang, Savarese, Ermon, et~al.]{zhang2023hive}
Shu Zhang, Xinyi Yang, Yihao Feng, Can Qin, Chia-Chih Chen, Ning Yu, Zeyuan Chen, Huan Wang, Silvio Savarese, Stefano Ermon, et~al.
\newblock Hive: Harnessing human feedback for instructional visual editing.
\newblock \emph{arXiv preprint arXiv:2303.09618}, 2023.

\end{thebibliography}
\bibliographystyle{neurips}
\newpage
\appendix
\section*{Appendix}
In the Appendix, Section \ref{app-sec-experiments} contains additional experiments and ablations, Section \ref{app-sec-qual} contains qualitative results and examples from different perceptual datasets, and Section~\ref{app-sec-details} contains methodological/implementation details.

\section{Experiments}\label{app-sec-experiments}
\subsection{Classification with the VTAB Benchmark}
\label{sup:vtab_classification}

We evaluate on the Visual Task Adaptation Benchmark (VTAB) \cite{Zhai2019ALS}, a comprehensive set of nineteen classification datasets that is commonly used to study the transferability of representations. VTAB covers a broad set of domains beyond classic natural image datasets, including medical images, scene understanding, and fine-grained classification. Several of the datasets are drawn from domains that are not included in the NIGHTS dataset.

In Tables \ref{tab:vtab-results}a-b, we show classification results on all VTAB datasets for each backbone, and its human-aligned version (denoted with -HA). The VTAB datasets are divided into three categories: Natural, Specialized, and Structured. Natural datasets include standard vision benchmarks; specialized datasets include benchmarks captured through specialized imaging equipment, such as satellites and microscopes; structured datasets include benchmarks that focus on structure and layout, with both naturally-captured and simulated images. 

We find that human-aligned models largely lead to worse performance on standard natural tasks. Results on specialized tasks are mixed; for most datasets, human alignment helps or hurts by a marginal amount. Results on structured datasets (Table \ref{tab:vtab_structure}) are also mixed, however on particular datasets such as sNORB, dSprites-Orientation, and dSprites-Location, alignment improves performance across backbones. We remark, however, that these structured datasets may have limited evaluation utility. The Clevr-Count task, for instance, has a numerical ground truth (i.e. the number of objects in an image) in which the distance from the correct answer is meaningful; thus it may be ill-suited to classification, and better suited to continuous evaluations such as RMSE or MAE, which we report in Section \ref{sec:counting}.

\begin{table*}[ht]
\centering
\begin{minipage}{0.66\linewidth}
    \centering
    \resizebox{1.0\linewidth}{!}{
        \begin{tabular}{l|lllllll}
\multicolumn{1}{c}{} 
& \multicolumn{7}{c}{\bf Natural} \\
 {\bf Model}
 & \begin{sideways}{Caltech101}\end{sideways} 
 & \begin{sideways}{CIFAR-100}\end{sideways} 
 & \begin{sideways}{DTD}\end{sideways} 
 & \begin{sideways}{Flowers102}\end{sideways} 
 & \begin{sideways}{Pets}\end{sideways} 
 & \begin{sideways}{Sun397}\end{sideways} 
 & \begin{sideways}{SVHN}\end{sideways} \\ 
 \midrule
  DINO & \bf 95.3 & \bf 83.0 & \bf 78.7 & \bf 96.7 & \bf 93.1 & \bf 74.8 & 70.8 \\
  DINO-HA & 93.6 & 80.8 & 72.6 & 93.2 & 91.0 & 69.7 &  \bf 73.2 \\
 \midrule
  DINOv2 & 95.5 & \bf{88.9}$^{\dagger}$ & \bf 82.8 & \bf{99.7}$^{\dagger}$ & \bf{95.9}$^{\dagger}$ & \bf 81.9 & 61.2 \\
  DINOv2-HA & \bf 96.2 & 87.7 & 80.2 & 97.9 & 89.7 & 77.0 & \bf 63.3 \\
 \midrule
  CLIP & 95.4 & \bf 76.5 & \bf 76.6 & \bf 95.5 & \bf 87.6 & \bf 80.6 & 65.4 \\
  CLIP-HA & \bf 95.7 & 76.4 & 76.0 & 93.4 & 85.5 & 78.1 & \bf 69.2 \\
 \midrule
  OpenCLIP & \bf 97.3 & \bf 82.4 & \bf 81.6 & \bf 97.0 & \bf 90.4 & \bf 82.3 & \bf 73.1 \\
  OpenCLIP-HA & 96.0 & 75.8 & 77.3 & 93.5 &  83.7 & 76.6 & 70.9 \\
 \midrule
  SynCLR & \bf 95.5 & \bf 74.3 & \bf 80.5 & \bf 98.8 & \bf 92.9 & \bf 81.4 & 56.2 \\
  SynCLR-HA & 95.0 & 69.7 & 77.8 & 95.9 & 90.5 & 75.7 & \bf 70.9 \\
 \midrule
  Ensemble & \bf{97.4}$^{\dagger}$ & \bf 86.8 & \bf{84.3}$^{\dagger}$ & \bf 98.8 & \bf 94.4 & \bf{83.9}$^{\dagger}$ & 80.1  \\
  Ensemble-HA  & 97.3 & 85.0 & 81.4 & 96.4 & 91.9 & 82.5 & \bf{82.2}$^{\dagger}$ \\
 \bottomrule
\end{tabular}

    }
\end{minipage}%
\hspace{0.2cm} 
\begin{minipage}{0.295\linewidth}
    \centering
    \resizebox{1.0\linewidth}{!}{
        \begin{tabular}{llll}
\multicolumn{4}{c}{\bf Specialized} \\
 \begin{sideways}{Camelyon}\end{sideways} 
 & \begin{sideways}{EuroSAT}\end{sideways} 
 & \begin{sideways}{Resisc45}\end{sideways} 
 & \begin{sideways}{Retinopathy}\end{sideways}  \\ 
 \midrule
  85.8 & 97.2 & \bf 93.8 & \bf 77.9 \\
  83.7 & \bf 97.3 & 93.2 & 76.9 \\
 \midrule
  \bf{86.7}$^{\dagger}$ & 96.4 & 92.7 & \bf 78.6  \\
  84.6 & \bf 97.0 & \bf 93.5 & 77.9 \\
 \midrule
  83.7 & 95.3 & 92.9 & 75.8 \\
  \bf 84.0 & \bf 95.9 & \bf 93.0 & \bf 76.1 \\
 \midrule
  \bf 82.3 & 96.4 & \bf 83.3 & 96.1 \\
  82.9 & \bf 96.5 & 92.6 & \bf 76.2 \\
 \midrule
  \bf{86.0} & 96.6 & \bf 94.0 & \bf{78.8}$^{\dagger}$\\
  85.1 & \bf 96.8 & 93.4 & 78.0 \\
 \midrule
  85.7 & \bf{98.0}$^{\dagger}$ & \bf{96.1}$^{\dagger}$ & \bf 78.7 \\
  83.5 & 97.9 &\bf{96.1}$^{\dagger}$ & 78.1 \\
 \bottomrule
\end{tabular}

    }
\end{minipage}
\captionof{table}{\small {\textbf{Performance on VTAB natural subset (left) and specialized subset (right).}}}\label{tab:vtab-results}
\end{table*}

\begin{table*}[ht]\label{tab:vtab_results}
\begin{center}
\resizebox{0.6\linewidth}{!}{
\begin{tabular}{l|llllllll}
\multicolumn{1}{c}{} 
& \multicolumn{8}{c}{\bf Structured}\\
 {\bf Model} 
 & \begin{sideways}{Clevr-Count}\end{sideways} 
 & \begin{sideways}{Clevr-Dist}\end{sideways} 
 & \begin{sideways}{DMLab}\end{sideways} 
 & \begin{sideways}{dSpr-Loc}\end{sideways} 
 & \begin{sideways}{dSpr-Orti}\end{sideways} 
 & \begin{sideways}{KITTI-Dist}\end{sideways} 
 & \begin{sideways}{sNORB-Azim}\end{sideways} 
 & \begin{sideways}{sNORB-Elev}\end{sideways} \\ 
 \midrule
  DINO & 82.2 & \bf 58.1 & \bf 53.6 & 63.8 & 57.1 & \bf 72.6 & 61.8 & \bf 56.5 \\
  DINO-HA & 79.2 & 57.2 & 51.1 & \bf 81.5 & \bf 66.6 & 66.2 & \bf 65.8 & 55.3 \\
 \midrule
  DINOv2 & 87.1 & 54.6 & \bf 57.7 & 59.7 & 68.2 & 62.3 & 55.3 & 52.8  \\
  DINOv2-HA & 83.6 & \bf 59.0 & 53.7 & \bf 79.2 & \bf 74.1 & \bf 69.1 & \bf 64.7 & \bf 55.9 \\
 \midrule
  CLIP & 70.6 & 55.0 & 50.2 & 67.6 & 55.6 & \bf 65.3 & 37.5 & 47.0 \\
  CLIP-HA & \bf 72.9 & \bf 57.8 & \bf 51.4 & \bf 80.2 & \bf 60.3 & 64.6 & \bf 42.6 & \bf 47.2 \\
 \midrule
  OpenCLIP & \bf 78.5 & 53.8 & \bf 52.0 & 63.3 & 55.0 & \bf 67.9 & 38.0 & \bf 50.4  \\
  OpenCLIP-HA & 76.7 & \bf 57.6 & 51.5 & \bf 84.9 & \bf 64.1 & 63.3 & \bf 46.7 & 50.1 \\
 \midrule
  SynCLR  & \bf{90.9}$^{\dagger}$ & \bf {63.8}$^{\dagger}$ & \bf{57.3} & 68.1 & 55.2 & \bf{76.9}$^{\dagger}$ & 52.1 & \bf 60.5 \\
  SynCLR-HA & 81.7 & 61.5 & 52.9 & \bf 86.1 & \bf 64.3& 63.6 & \bf 66.6 & 58.5 \\
 \midrule
  Ensemble & \bf 89.2 & 63.5 & \bf{58.4}$^{\dagger}$ & 72.5 & 67.3 & \bf{76.5} & 76.9 & \bf{65.5}$^{\dagger}$ \\
  Ensemble-HA & 85.5 & \bf{63.8}$^{\dagger}$ & 57.1 & \bf{91.8}$^{\dagger}$ & \bf{76.4}$^{\dagger}$ & 70.0 & \bf{81.0}$^{\dagger}$ & 62.2\\
 \bottomrule
\end{tabular}
}
\end{center}
\caption{\small {\bf Performance on VTAB structured subset.}}\label{tab:vtab_structure}
\end{table*}

\subsection{Additional dataset ablations} \label{app-ablation}
\begin{figure*}[ht!]
    \centering
    \includegraphics[width=\linewidth]{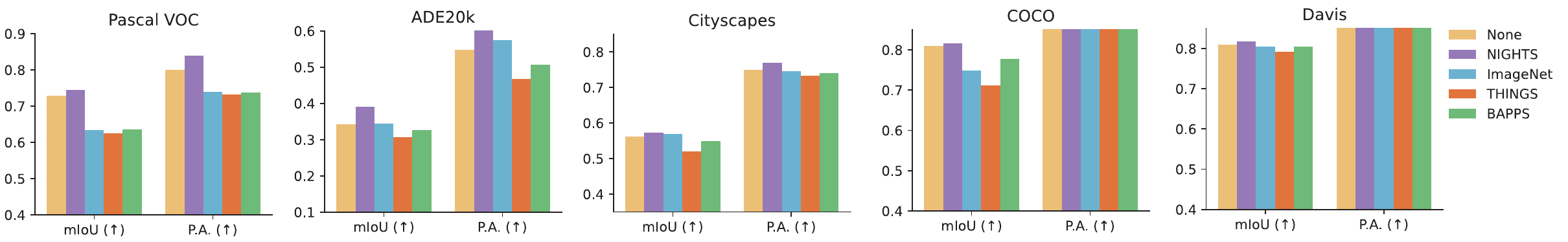}
    \caption{\small \textbf{Dataset ablations on semantic segmentation. } Following the same procedure as in section 4.5 and the segmentation training in section 4.1 of the main paper, we ablate low/mid/high-level similarity. We find that tuning on mid-level similarity with NIGHTS provides the most improvement, with many other cases degrading the base DINO backbone.}
    \label{fig:app-ablation-seg}
\end{figure*}

\begin{figure*}[ht!]
    \centering
    \includegraphics[width=0.7\linewidth]{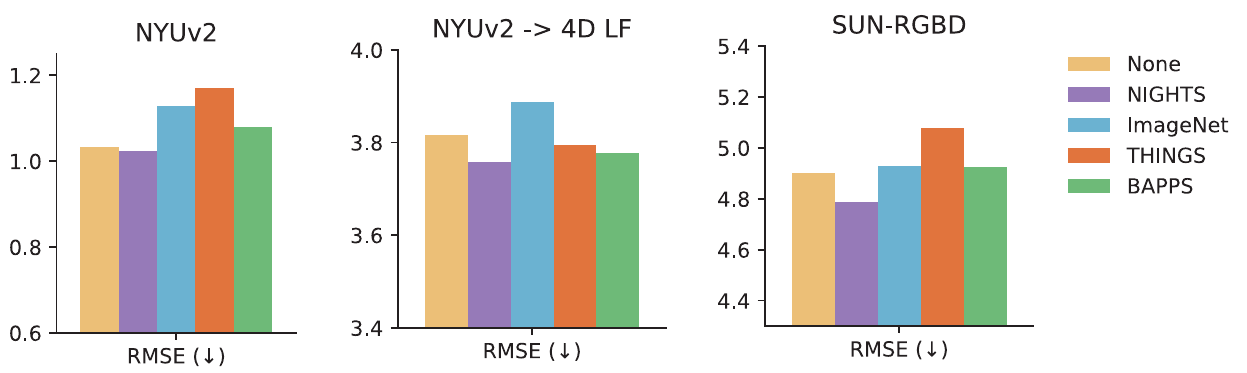}
    \caption{\small \textbf{Dataset ablations on depth estimation. } Following the ablation and depth training setups described in the main paper, we evaluate low/mid/high-level similarity. We find that tuning on mid-level similarity with NIGHTS results in the lowest error (RMSE), and that the other cases often result in worse performance than base DINO.}
    \label{fig:app-ablation-dense}
\end{figure*}

In Section \ref{sec:ablation} we show that tuning on NIGHTS leads to larger performance improvements on object counting and instance retrieval than training on other triplet datasets. Here, we show that this finding is consistent across dense prediction tasks as well. We evaluate models trained on NIGHTS, BAPPS, THINGS, and ImageNet -- as described in section \ref{sec:ablation} -- on semantic segmentation and depth estimation. As shown in Fig.~\ref{fig:app-ablation-seg}-~\ref{fig:app-ablation-dense}, training on NIGHTS outperforms training on all other datasets.

\begin{figure*}[ht!]
    \centering
    \includegraphics[width=1.0\linewidth]{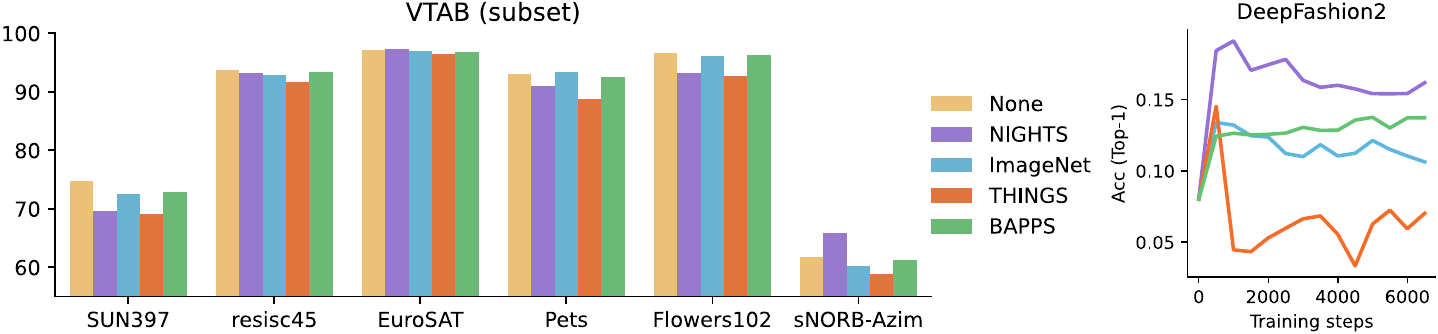}
    \caption{\small \textbf{(left) Dataset ablations on a subset of VTAB. } Following the same procedure as in section 4.5 and the segmentation training in section 4.1 of the main paper, we ablate low/mid/high-level similarity. We find that all perceptual datasets degrade the DINO backbone, with the exception of sNORB, for which NIGHTS is best. We also note that training on BAPPS or ImageNet often better preserves base model performance than NIGHTS. \textbf{(right) Training-step ablations on DeepFashion2. } We ablate low/mid/high-level similarity and test models finetuned for different numbers of steps on instance retrieval. We find that a small amount of finetuning on NIGHTS is best, after which performances declines (though remains above the base model). \textit{Note that the legend applies to both the left and right figure.}}
    \label{fig:app-ablation-vtab}
\end{figure*}

We additionally run this ablation on a subset of VTAB (see Fig.~\ref{fig:app-ablation-vtab}a). On these classification datasets, base models perform best; the exception is sNORB (pose prediction) for which NIGHTS is best. Amongst perceptual datasets, NIGHTS is sometimes outperformed by BAPPS/ImageNet. This result is consistent with our findings in Section \ref{sup:vtab_classification}, that tuning on NIGHTS often fails to improve classification performance.

Finally, we ablate the \textit{strength} of alignment -- i.e. the training loss when fine-tuning on different perceptual datasets. In Fig.~\ref{fig:app-ablation-vtab}b we show the downstream performance on DeepFashion2 when fine-tuning for increasing numbers of steps on different datasets. Tuning on NIGHTS outperforms other datasets over the full training trajectory. Moreover, while performance rises significantly with a small amount of alignment to NIGHTS, it trends down after >1000 steps. This indicates that a small amount of alignment is helpful, however overfitting may harm performance.

\subsection{Additional RAG results}
\begin{figure*}[ht!]
    \centering
    \includegraphics[width=1.0\linewidth]{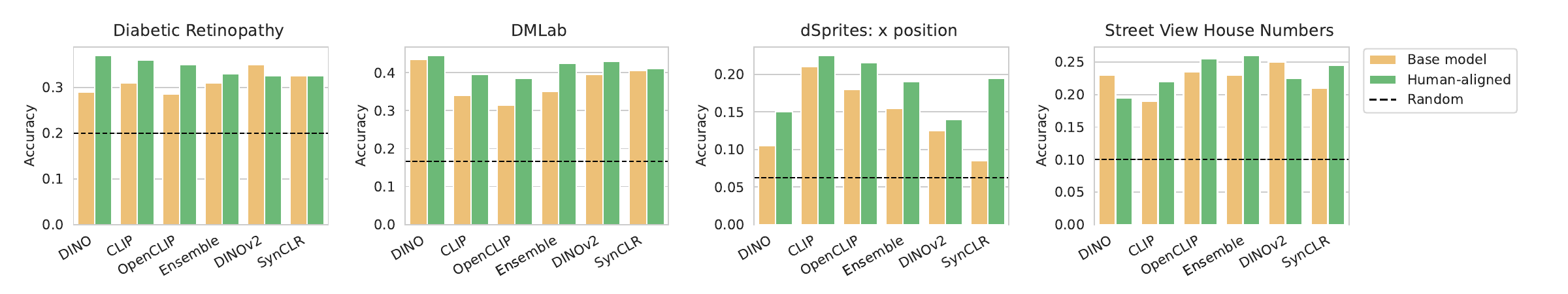}
    \caption{\small \textbf{Retrieval-augmented generation results on image classification for four different VTAB datasets.} We evaluate the IDEFICS2 model, an 8B-parameter multimodal model released more recently than OpenFlamingo and capable of in-context learning. We follow the same experimental setup in retrieving examples as detailed in \ref{sec:rag}.}
    \label{fig:app-rag}
\end{figure*}

We replicate our RAG experiment on OpenFlamingo in Section \ref{sec:rag} on IDEFICS2, a recently released 8B multimodal model achieving state-of-the-art results across several benchmarks \cite{Tong_2024_CVPR}. As shown in Fig. \ref{fig:app-rag}, across the same four classification datasets, performance consistently improved when using NIGHTS-tuned models in the RAG pipeline. These results validate our RAG results on OpenFlamingo, suggesting that performance gains from perceptual alignment are not specific to a particular VLM.

\subsection{Perceptual alignment for CNNs}
Our main experiments assess the effects of perceptual alignment for an array of ViT models, as these are state-of-the-art pretrained vision representations. Of further interest, however, is whether the same effects are observed for pre-trained CNNs as well.

We assess the effects of perceptual alignment with NIGHTS on two popular CNNs: ResNet50 and ConvNeXt-B, using the same loss described in Section \ref{subsec:objective}. We evaluate on counting and instance retrieval tasks; our results are shown in Tables \ref{tab:cnn_results}a-b. Similarly to ViTs, human-aligned CNNs outperform their pretrained counterparts across both tasks. Note that we report results for both ConvNeXt and ResNet, however acknowledge that the accuracies for ResNet are likely too small to draw conclusions from.

\begin{table*}[ht]
\centering
\begin{minipage}{0.39\linewidth}
    \centering
    \resizebox{1.0\linewidth}{!}{
        \resizebox{1.0\linewidth}{!}{
\begin{tabular}{lcc}
\toprule
 & \multicolumn{2}{c}{\bf Clevr-Count} \\
 {\bf Model} & {RMSE} & {MAE} \\
 \midrule
  ConvNeXt       & 2.045         & 1.522        \\
  ConvNeXt-HA    & \bf 1.631     & \bf 1.193        \\
 \midrule
  ResNet         & 3.140         & 2.551        \\
  ResNet-HA      & \bf 1.729     & \bf 1.282    \\
 \bottomrule
\end{tabular}
}

    }
\end{minipage}%
\hspace{0.1cm} 
\begin{minipage}{0.47\linewidth}
    \centering
    \resizebox{1.0\linewidth}{!}{
        \resizebox{1.0\linewidth}{!}{
\begin{tabular}{lccc}
\toprule
 & \multicolumn{3}{c}{\bf DeepFashion2} \\
 {\bf Model} & {Top-1} & {Top-3} & {Top-5} \\
 \midrule
  ConvNeXt         & 2.12           & 3.56           & 4.59           \\
  ConvNeXt-HA      & \bf 2.81       & \bf 4.8        & \bf 5.98       \\
 \midrule
  ResNet           & 0.018          & \bf 0.12       & 0.14           \\
  ResNet-HA        & 0.018          & 0.074          & \bf 0.17       \\
 \bottomrule
\end{tabular}
}

    }
\end{minipage}
\caption{\small {Human-aligned vs. pretrained CNNs, evaluated on counting and instance-retrieval. In most cases, human-aligned CNNs perform better.}}\label{tab:cnn_results}
\end{table*}

Due to a lack of open-source LoRA implementations for CNNs, we instead train MLPs on top of the respective final layers. Previous work \cite{dreamsim} indicated that training MLPs with NIGHTS can still affect representations, but may be less effective than full fine-tuning. Nonetheless, we still see downstream improvements with this method.

\section{Qualitative examples}\label{app-sec-qual}
\subsection{Additional visualizations}
\begin{figure*}[ht]
    \centering
    \includegraphics[width=\linewidth]{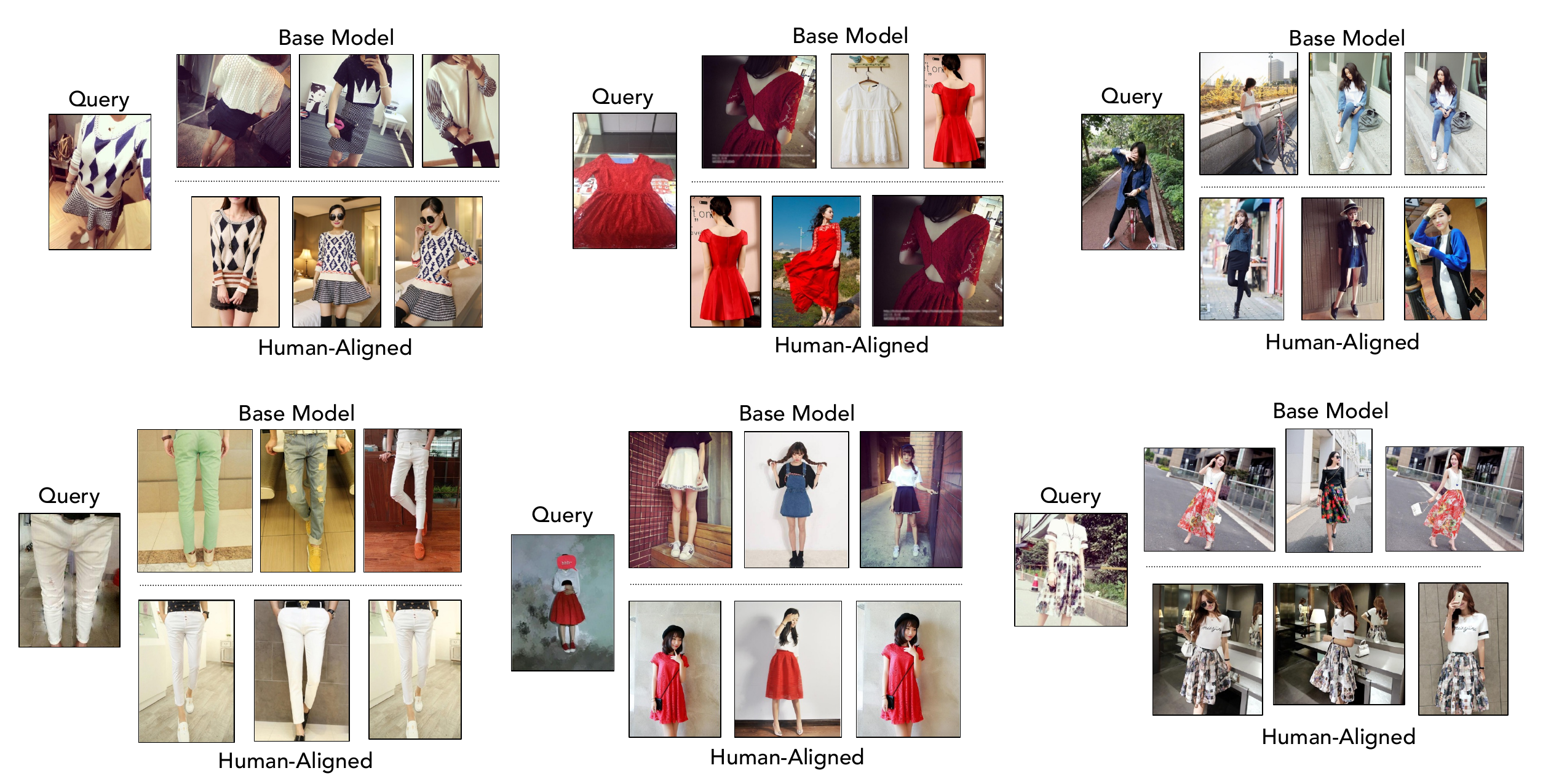}
    \caption{\small Additional visualizations for top-3 retrieved examples for a given query image in DeepFashion2. }
    \label{fig:fashion-supp}
\end{figure*}

\begin{figure*}[ht]
    \centering
    \includegraphics[width=\linewidth]{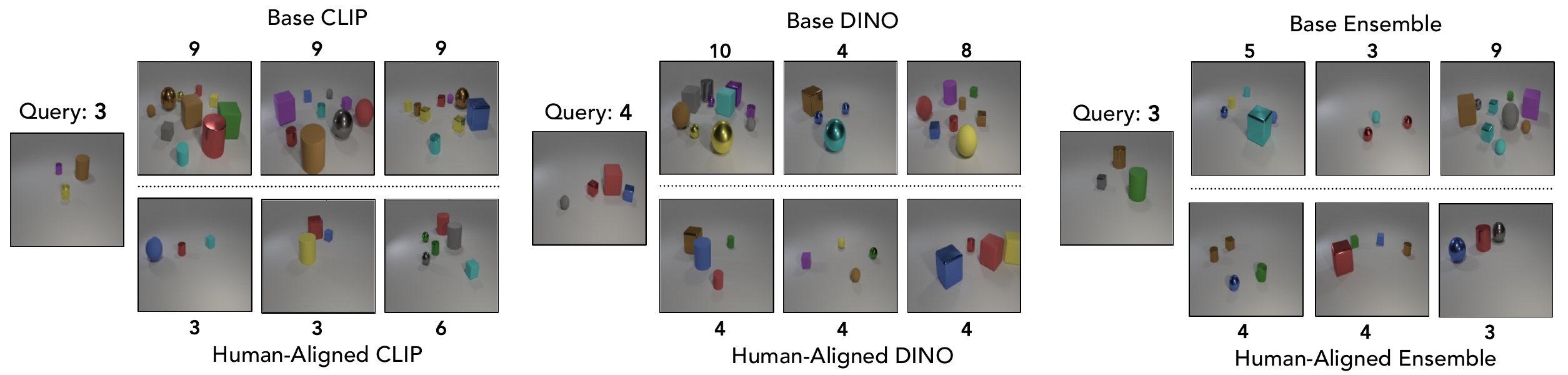}
    \caption{\small Additional Clevr-count examples comparing base and human-model nearest-neighbor image retrievals. }
    \label{fig:counting-supp}
\end{figure*}

See Figures \ref{fig:fashion-supp} and \ref{fig:counting-supp} for additional examples of instance retrieval on DeepFashion2 and object counting on Clevr-count.

\subsection{Dataset examples}
\begin{figure*}[ht!]
    \centering
    \includegraphics[width=\linewidth]{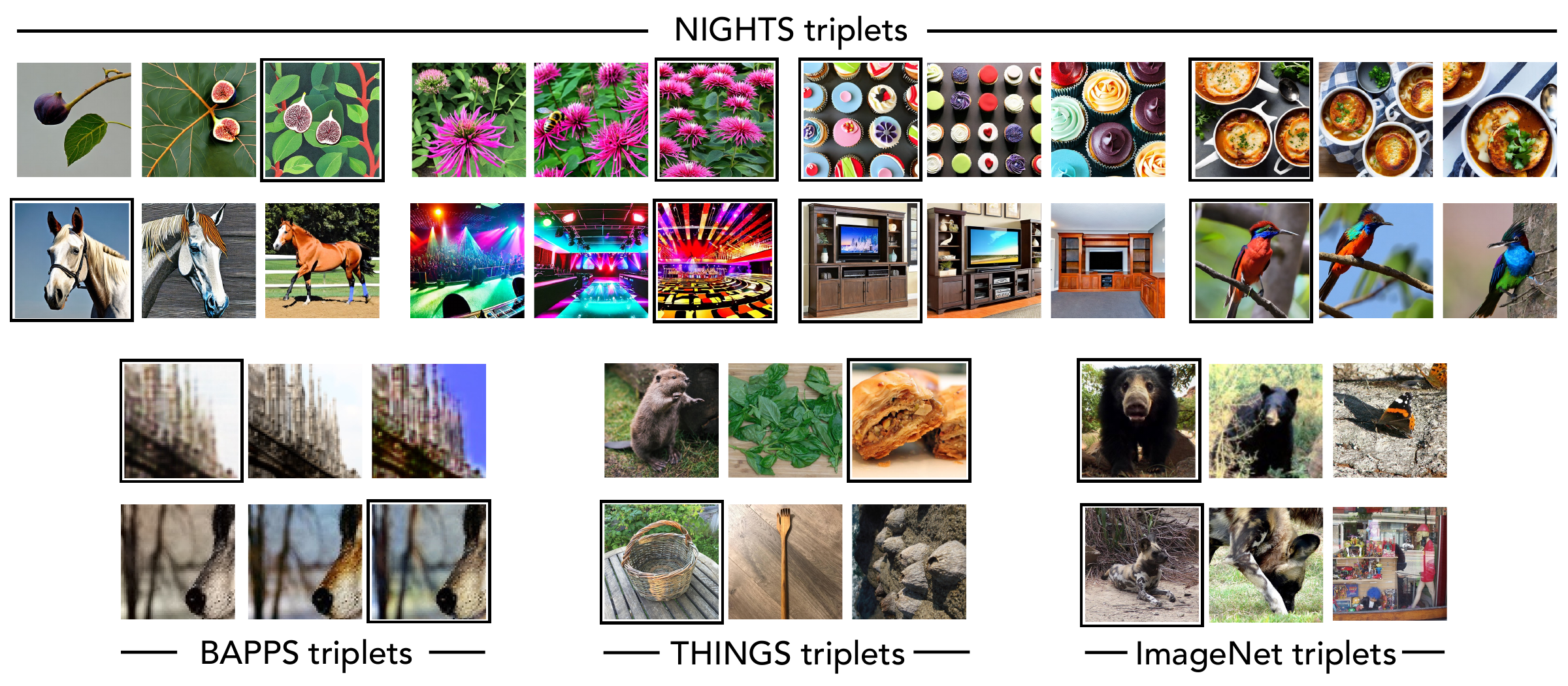}
    \caption{\small Examples of triplets from the NIGHTS, BAPPS, THINGS, and ImageNet datasets, with the bordered images labeled as more similar to the reference (middle image in each triplet). }
    \label{fig:dataset}
\end{figure*}
See Figure \ref{fig:dataset} for examples of the triplet datasets we train on in this paper. In each triplet, the reference image (middle) and outlined image are labeled as the similar pair. Perceptually-aligned models in this paper are tuned on NIGHTS triplets (top row), whose image variations encompass a variety of mid-level perceptual attributes including object count, identity, layout, subject pose, and color. In contrast, the datasets studied in our ablation encompass differing image attributes: BAPPS focuses on image patches with low-level distortions such as CNN artifacts and color jitter, THINGS encodes similarity in concept space, and our constructed ImageNet triplets outline class boundaries.

\section{Implementation Details}\label{app-sec-details}
\subsection{Training human-aligned models}
\label{ha-details}
The human-aligned models studied in this paper were fine-tuned with LoRA parameters: $r=16$, $\alpha=0.5$, $p=0.0$. All contrastive training (for image- and patch-level objectives) are trained with an Adam optimizer ($lr=0.0003$), batch size of 16, and a hinge loss margin of $m=0.05$ as detailed in \cite{dreamsim}. We train all models for 8 epochs and evaluate on the checkpoint with the lowest validation loss. Train/val/test splits on NIGHTS, BAPPS, and THINGS were used as-provided in the dataset.

\subsubsection{Dense prediction heads}
Each semantic segmentation head is a single linear layer mapping frozen DINO/DINOv2 patch tokens to a segmentation map. We train with the Adam optimizer ($\alpha=0.0003$) for 10 epochs with a batch size of 16 and 4 workers. The semantic segmentation head is optimized with a Jaccard Index loss (equivalently, mean Intersection-over-Union) defined as $J(A,B) = \frac{|A\cap B|}{|A\cup B|}$. 

Dense prediction heads are defined as a single linear layer mapping frozen DINO/DINOv2 patch tokens to a spatial map, whose values are then categorized into 256 bins. We train with the Adam optimizer ($\alpha=0.0003$) for 10 epochs with a batch size of 128 and 12 workers. The scale-invariant log loss for prediction $A$ and target $B$ is defined as: 
$$\mathcal{L}(A, B) = \frac{1}{N}\sum\limits_i d_i^2 + 0.15 \cdot \frac{1}{N^2} (\sum\limits_i d_i)^2$$
where $d_i = \log(A + \epsilon) - \log(B + \epsilon)$, $\epsilon=0.001$ to avoid gradient issues, and $N$ is the number of valid pixels.

All training and evaluation for dense prediction tasks is done on a single NVIDIA Titan RTX GPU. 

\subsection{Retrieval-augmented generation}
We follow the evaluation protocol in the codebase provided by \cite{awadalla2023openflamingo}, constructing a prompt containing three image/text examples, a query image, and a candidate class. We then average the model's output logits, and repeat for each candidate class in a dataset to form a softmaxed array of log probabilities. OpenFlamingo's classification prediction is determined as the class with the highest corresponding log probability, while classifications from IDEFICS2 are taken purely from a multiple-choice text response. Error bars were calculated as 95\% confidence intervals over five trials with 1,000 randomly-sampled queries each. 
All evaluation for RAG tasks is done on a single NVIDIA Titan RTX GPU. 
\subsubsection{Object counting}
Our k-Nearest Neighbor evaluations reported in the Counting section are run over $k\in \{1, 3, 5, 10\}$ for the training set, and the argmax $k$ with highest classification accuracy is used to report the final performance on the test set. We use the train and test sets as provided for the FSC147 and CARPK datasets, and construct a random 80:20 data split for the Clevr-Count dataset. 
All evaluation for this section is done on a single NVIDIA Titan RTX GPU. 

\subsection{Instance retrieval}
We evaluate backbones on instance retrieval by extracting features (CLIP/OpenCLIP: embedding, DINO/DINOv2/SynCLR: CLS token) from the gallery and query sets. For every query image, we compute the cosine similarity with every image in the gallery. We then calculate top-$k$ accuracy as the number of queries for which a correct gallery pair image was in the top $k$ retrievals.

\subsection{Dataset ablations}
We train models for the dataset ablation section with identical hyperparameters as the human-aligned models (details in Section \ref{ha-details} and on the same amount of data as the NIGHTS training set (13,900 triplets). We randomly select these triplets without replacement from the BAPPS and THINGS training sets, and construct the ImageNet triplets by randomly sampling two images from one class and a third image from a different class.

\subsection{VTAB classification}
To evaluate models on VTAB, we follow standard procedure and train a linear classifier on top of the frozen representations using multinomial logistic regression, with the sci-kit learn implementation \cite{muttenthaler2023improving}. For each classifier we perform a hyperparameter search over the regularization strength using 10-fold cross validation over the training set, and report performance on the validation set. We search over the values $c = \{1e0, 1e1, 1e2, 1e3, 1e4, 1e5, 1e6\}$ where $c$ is the inverse regularization strength.

\subsection{Additional compute details}
\label{compute-details}
See subsections above for experiment-specific compute details. This full research project required additional compute for experiments and results that are not included in this paper; these computations were also done on single NVIDIA Titan RTX, GeForce 2080, GeForce 3090, and V100 GPUs.

\subsection{Licenses for existing assets}
The datasets and models we use are released under the following licenses: \\
\begin{center}
\begin{minipage}{0.35\textwidth}
\begin{tabular}{@{}cc@{}}
\toprule
\multicolumn{1}{c}{\textbf{Dataset}} & \multicolumn{1}{c}{\textbf{License}} \\ \midrule
NIGHTS & MIT \\
VTAB & Apache 2.0 \\
BAPPS & BSD 2-Clause \\
THINGS & CC0 1.0 Universal \\
ImageNet & CC BY-NC \\
FSC147 & Apache 2.0 \\
CARPK & CC0 1.0 Universal \\
Pascal VOC & CC BY 2.0 \\
ADE20k & BSD 3-Clause \\
Cityscapes & CC BY-NC \\
COCO & CC BY 4.0 \\
DAVIS2017 & CC BY 4.0 \\
NYUv2 & MIT \\
4D Light Field & CC BY-NC-SA 4.0 \\
SUN-RGBD & MIT \\
DeepFashion & CC BY-NC \\ \bottomrule
\end{tabular}
\end{minipage}
\hfill
\begin{minipage}{0.4\textwidth}
\begin{tabular}{@{}cc@{}}
\toprule
\multicolumn{1}{c}{\textbf{Model}} & \multicolumn{1}{c}{\textbf{License}} \\ \midrule
DINO & Apache 2.0 \\
CLIP & MIT \\
OpenCLIP & MIT \\
DreamSim & MIT \\
SynCLR & Apache 2.0 \\
\bottomrule
\end{tabular}
\end{minipage}
\end{center}

\end{document}